\documentclass[10pt,twocolumn,letterpaper]{article}

\usepackage[pagenumbers]{cvpr} 

\usepackage{multirow, multicol}
\usepackage[utf8]{inputenc}
\usepackage{pifont}
\usepackage{booktabs}
\usepackage{colortbl}
\usepackage{xcolor}
\usepackage[table]{xcolor}

\usepackage{cuted}
\usepackage{capt-of}
\usepackage{tikz}
\usepackage{gensymb}
\usepackage{arydshln}

\newcommand{\xmark}{\textcolor{red}{\ding{55}}}%
\newcommand{\cmark}{\textcolor{ForestGreen}{\ding{51}}}%

\newcommand{\best}[1]{\cellcolor[HTML]{b8e0bc}\textbf{#1}} 
\newcommand{\second}[1]{\cellcolor[HTML]{fdf6b4}#1} 
\newcommand{\third}[1]{\cellcolor[HTML]{ffdcc0}#1} 
\newcommand{\grayrow}{\rowcolor[gray]{0.95}} 

\definecolor{cvprblue}{rgb}{0.21,0.49,0.74}
\usepackage[pagebackref,breaklinks,colorlinks,allcolors=cvprblue]{hyperref}

\def\paperID{} 
\def\confName{CVPR}
\def\confYear{2026}

\title{VarSplat: Uncertainty-aware 3D Gaussian Splatting for Robust RGB-D SLAM}

\author{Anh Thuan Tran, Jana Košecká\\
Department of Computer Science\\
George Mason University\\
{\tt\small \{atran92, kosecka\}@gmu.edu}\\
\\
\href{https://anhthuan1999.github.io/varsplat/}{https://anhthuan1999.github.io/varsplat/}
\vspace{-50pt}
}

\begin{document}

\maketitle

\begin{strip} \label{fig1}
\centering
\includegraphics[width=\textwidth]{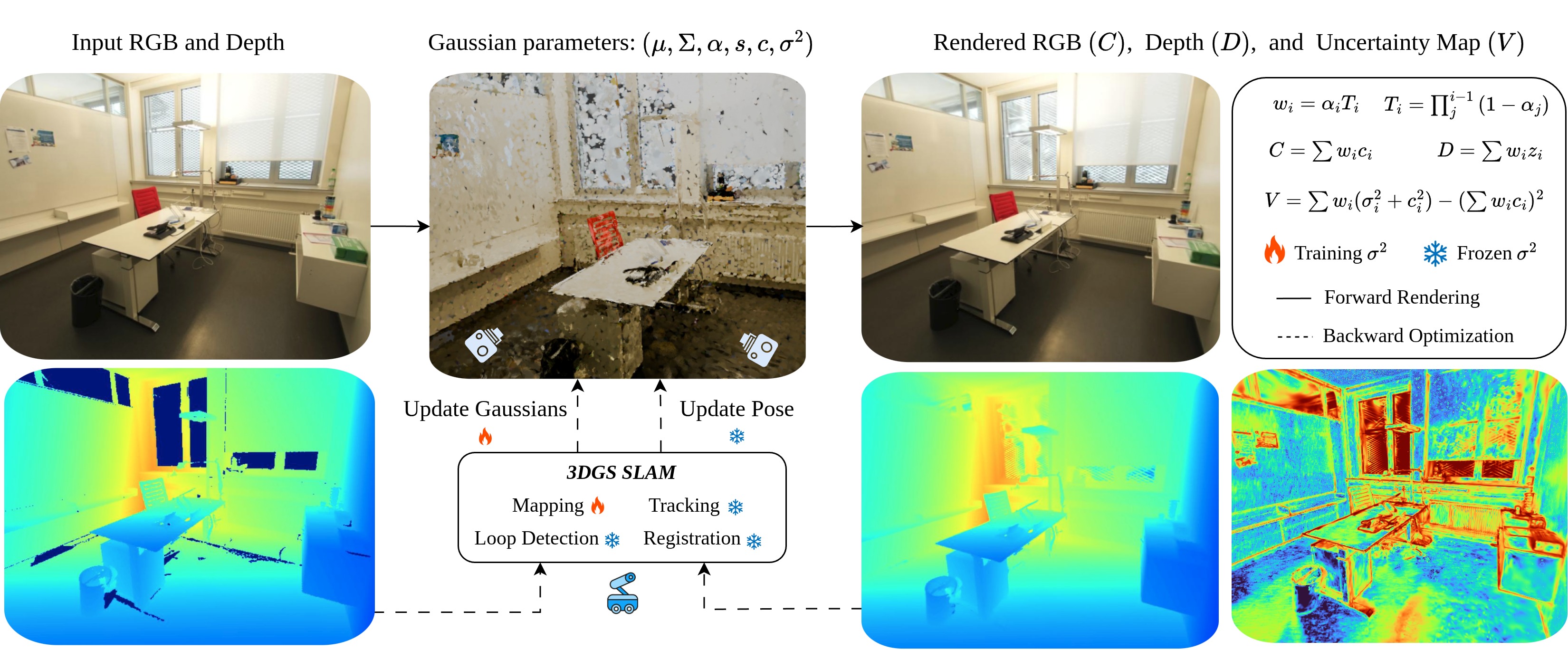} 
\captionof{figure}{\textbf{VarSplat.}
Given RGB-D inputs, each 3D Gaussian jointly learns position, orientation, scale, color, opacity, and appearance variance $\sigma^2$.
During \textcolor{orange!80!black}{mapping}, $\sigma^2$ is optimized jointly with other Gaussian parameters.
The rendered per-pixel uncertainty $V$ serves as a confidence weight during
\textcolor{cyan!80!black}{tracking} and \textcolor{cyan!80!black}{registration}, while the per-Gaussian variance $\sigma^2$ further guides \textcolor{cyan!80!black}{loop detection}, enabling robust and uncertainty-aware SLAM.
\label{fig:overview}}
\end{strip}

\begin{abstract}
Simultaneous Localization and Mapping (SLAM) with 3D Gaussian Splatting (3DGS) enables fast, differentiable rendering and high-fidelity reconstruction across diverse real-world scenes.
However, existing 3DGS-SLAM approaches handle measurement reliability implicitly, making pose estimation and global alignment susceptible to drift in low-texture regions, transparent surfaces, or areas with complex reflectance properties. To this end, we introduce VarSplat, an uncertainty-aware 3DGS-SLAM system that explicitly learns per-splat appearance variance. By using the law of total variance with alpha compositing, we then render differentiable per-pixel uncertainty map via efficient, single-pass rasterization. This map guides tracking, submap registration, and loop detection toward focusing on reliable regions and contributes to more stable optimization. 
Experimental results on Replica (synthetic) and TUM-RGBD, ScanNet, and ScanNet++ (real-world) show that VarSplat improves robustness and achieves competitive or superior tracking, mapping, and novel view synthesis rendering compared to existing studies for dense RGB-D SLAM.
\end{abstract}

\section{Introduction}
\label{sec:intro}

Simultaneous Localization and Mapping (SLAM) \cite{denseslam} \textcolor{blue} is significant for robotics and AR/VR applications by {giving the autonomous systems ability to actively explore unknown environments while incrementally building a map. }
Classical methods \cite{dtam,kinectfusion,elasticfusion} based on voxels, TSDFs, surfels, or meshes can achieve real-time tracking and mapping but struggle to capture photorealistic appearance. To improve visual fidelity, Neural radiance fields (NeRF) \cite{nerf} have been integrated into SLAM systems \cite{imap,niceslam}, though implicit volumetric rendering remains costly for real-time performance. 3D Gaussian Splatting (3DGS) \cite{3dgs} further addresses this bottleneck by rasterizing anisotropic Gaussians with alpha compositing, enabling fast, differentiable rendering. Recent 3DGS-SLAM systems \cite{splatam, gsslam, gaussianslam, loopsplat} adopt this explicit representation and jointly optimize camera poses and Gaussian parameters.

Despite these advances, a key limitation exists: \emph{measurement reliability is rarely modeled explicitly}. Uniform photometric weighting leaves pose estimation vulnerable in low-texture regions, around depth discontinuities, and on reflective surfaces. For safety-critical systems, explicitly quantifying uncertainty is important, providing confidence alongside rendering quality.
Prior efforts address uncertainty primarily on the {geometric} side 
(e.g., depth variance or probabilistic filters)
\cite{cgslam,uncleslam,unislam} or rely on {pretrained} predictors \cite{wildgsslam}. The {appearance} uncertainty that directly reflects instability in 3DGS rendering has not been treated as a first-class quantity in online dense SLAM.

To address these challenges, we introduce \textbf{VarSplat}, an uncertainty-aware RGB-D SLAM system leveraging 3D Gaussian Splatting. 
VarSplat integrates explicit uncertainty into the map representation by learning per-splat appearance variance $\sigma^{2}$ that is trained along with standard Spherical Harmonics (SH) coeffcients.
Using the law of total variance, we derive propagation through 3DGS rasterizer to obtain differentiable {per-pixel} uncertainty map $V$.
The system is trained end-to-end {online},
jointly optimizing poses, Gaussian parameters, and $\sigma^{2}$ as the map grows.
For downstream tasks, uncertainty is applied across all three stages. In \emph{tracking}, the per-pixel uncertainty map $V$ provides short-horizon reliability {measure} within each submap, improving frame-to-frame pose updates. In \emph{registration}, the same per-pixel reliability supports mid-range alignment between overlapping submaps. For \emph{loop detection}, per-splat variances 
$\sigma^{2}$ modulate submap similarity and correct long-range drift. 
To the best of our knowledge, VarSplat is the first 3DGS–SLAM system to learn per-splat appearance variance and render per-pixel uncertainty in online setting.
In summary, our {contributions} can be shown as follows:
\begin{itemize}
\item We introduce \textit{VarSplat}, an RGB-D 3DGS–SLAM system that learns per-splat appearance variance $\sigma^{2}$ to render differentiable per-pixel uncertainty map $V$ while maintaining single-pass rasterization efficiency.
\item We integrate uncertainty at both {representation} and {renderer} levels: poses, Gaussian parameters, and variance $\sigma^{2}$ are optimized together in a fully online, end-to-end submap pipeline.

\item Extensive experiments on Replica, TUM-RGBD, ScanNet, ScanNet++ show improved robustness and competitive or superior tracking, reconstruction, and novel view synthesis rendering against 3DGS-SLAM baselines.
\end{itemize}

\section{Related Work}
In this section, we briefly go through various approaches to dense SLAM that leverage radiance fields for 3D representations and uncertainty quantification. 
\\
\noindent
{\bf Dense Visual SLAM.} 
Traditional dense SLAM, with DTAM \cite{dtam} and KinectFusion \cite{kinectfusion}, achieved real-time tracking and mapping from RGB-D images using explicit volumetric or surface representations such as TSDFs \cite{bundlefusion,volumetricfusion}, voxels \cite{voxfusion,difusion} and surfels \cite{badslam,elasticfusion}. While effective for real-time pose estimation, these approaches struggle to provide texture-rich, high-fidelity reconstructions and synthesize novel views. 
iMap \cite{imap} integrated Neural Radiance Fields (NeRF) \cite{nerf} into tracking and mapping, and Nice-SLAM \cite{niceslam} improved scalability with multi-resolution feature grids.
Additional efficiency representations and global consistency concerns were addressed 
in~\cite{eslam,coslam,goslam,loopy,pointslam,pointnerf}.
Recently, SplaTAM \cite{splatam} used 3D Gaussians \cite{3dgs} for dense RGB-D SLAM with per-pixel photometric losses, and submap-based systems \cite{monogs,gsslam,gaussianslam} balanced accuracy and efficiency through incremental mapping. 
VarSplat follows this general 3DGS-SLAM architecture but treats measurement reliability explicitly. 
\\
\noindent
{\bf Uncertainty Quantification for Radiance Fields.} 
In Neural Radiance Fields, Stochastic NeRF \cite{stochasticnerf} and Conditional-Flow NeRF \cite{conditionalflownerf} attempted to quantify uncertainty by reparametrizing NeRF with Bayesian model. However, these approaches focused on predictive uncertainty and do not capture parameter uncertainty grounded in observed information.
Bayes’ Rays \cite{bayesnerf} introduced spatially parameterized perturbation field and interpolates uncertainty at NeRF query points, estimating it with Laplace approximation.
Information-driven studies used uncertainty for decision making. FisherRF \cite{fisherrf} 
modeled uncertainty with Fisher information for active view selection. 
Bayesian NeRF \cite{bayesiannerf} modeled explicitly quantifies uncertainty in the geometric volume structure, particularly volume density and occupancy. 
ActiveNeRF \cite{actnerf} rendered per-pixel variance map using the neural network to guide model acquisition. 
Recent methods \cite{modelinguncertaintygs,variationalgs} integrated stochasticity into 3D Gaussian Splatting \cite{3dgs} pipeline through variational inference and Monte Carlo sampling to approximate pixel-level variance. A probabilistic view also treats 3DGS optimization as SGLD sampling \cite{3dgsmcmc}.
Sensitivity-based \cite{pup3dgs} approach related uncertainty to model sensitivity for pruning. Semantic 3DGS \cite{uncertaintysemanticgs} estimates per-pixel variance while rasterizing semantic maps.
VarSplat introduces a conceptually different method that learns per-splat appearance variance and leverage Gaussian rasterizer to render per-pixel uncertainty map.  This approach allows us to maintain single-pass rasterization efficiency in online setting.
\\
\noindent
{\bf Uncertainty in SLAM.} 
Incorporating uncertainty in SLAM is important for robust localization and consistent maps under real-time constraints. 
REMODE \cite{remode} models per-pixel depth distributions, then uses their variance into mapping, while SVO \cite{svo} provides depth filters with explicit uncertainty updates. DeepFactors \cite{deepfactors} presents a real-time probabilistic framework for dense monocular SLAM.
Within neural SLAM, UncLe-SLAM \cite{uncleslam} learns per-pixel depth uncertainty and reweights tracking and mapping. Uni-SLAM \cite{unislam} defines predictive uncertainty for neural implicit fields and uses it to reweight pixel losses. CG-SLAM \cite{cgslam} builds 3D Gaussian field and models depth uncertainty to guide optimization and select informative Gaussians. For dynamic scenes, WildGS-SLAM \cite{wildgsslam} predicts an uncertainty map from DINOv2 features to filter moving distractors, and GLC-SLAM \cite{glcslam} uses uncertainty for keyframe selection to stabilize submap optimization.
Rather than modeling geometric depth variance \cite{cgslam,uncleslam} or relying on pretrained predictors \cite{wildgsslam}, we learn per-splat appearance variance and propagate it through alpha compositing and the law of total variance to obtain a differentiable per-pixel uncertainty map. In contrast to variance from ray-termination probabilities \cite{unislam}, our variance is learned from the rasterizer and updated frame by frame.
\begin{figure*}
  \centering
  \includegraphics[width=\textwidth]{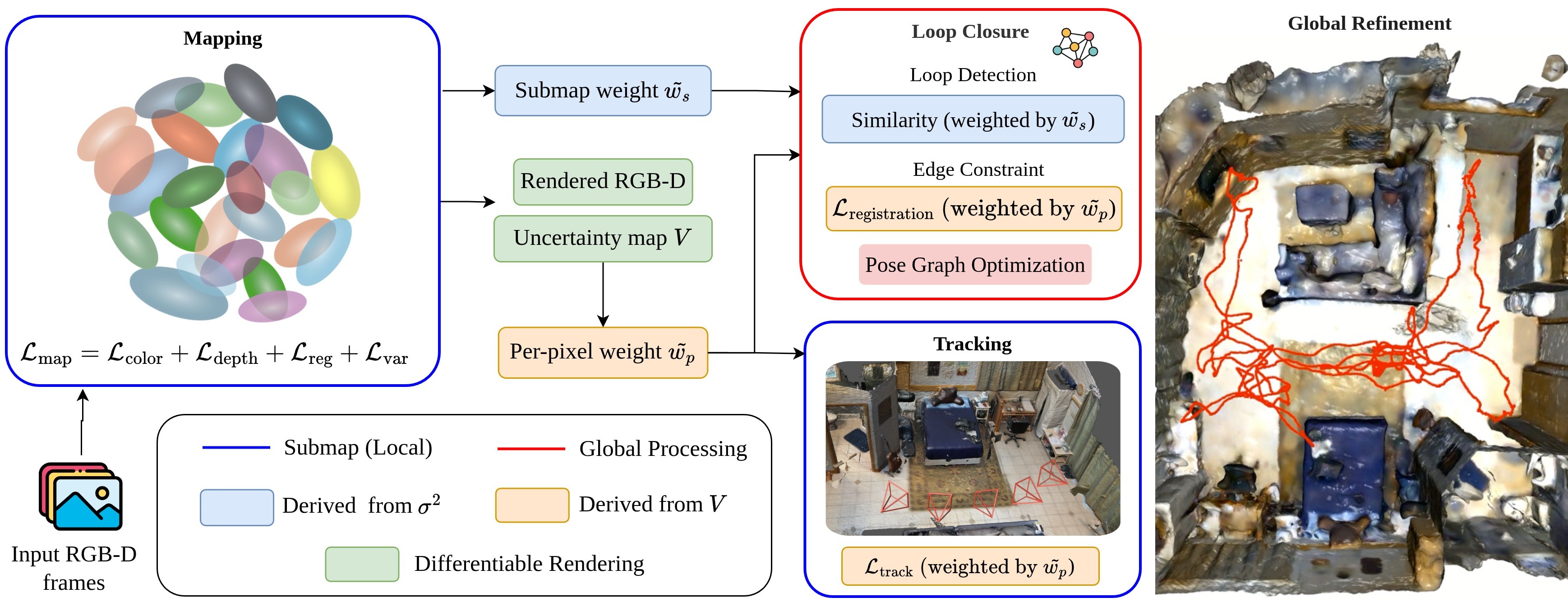}
  \caption{VarSplat architecture. During mapping, each 3D Gaussian jointly learns position, appearance, and variance \(\sigma^2\). The per-splat variances are composited into per-pixel uncertainty \(V\) through the differentiable rasterizer in \textcolor{blue}{mapping}. This uncertainty map is then served as confidence in \textcolor{blue}{tracking} and \textcolor{red}{registration}, while \(\sigma^2\) is used for uncertainty-aware \textcolor{red}{loop detection}.}
  \label{fig:short}
\end{figure*}
\section{Method}
VarSplat is an RGB-D SLAM approach that jointly estimates camera poses and incrementally updates 3D Gaussian Splatting (3DGS) map from input frames, following the general pipeline of \cite{gaussianslam, loopsplat}. However, pose estimation through photometric optimization can suffer from unreliable observations in low-texture regions, reflective surfaces, and areas near depth discontinuities, which can destabilize this process and potential drift.
To address these issues, we introduce a novel uncertainty quantification pipeline based on per-pixel uncertainty map rendering. 
Specifically, differentiable uncertainty map $V$ is rendered through 3DGS rasterizer via the law of total variance and alpha compositing using per-splat variance appearance \(\sigma^2\). Unlike previous methods relying on pretrained predictors \cite{wildgaussians, wildgsslam}, VarSplat learns this variance parameter from scratch, jointly optimizing them with appearance and geometry during mapping. 
The resulting \(\sigma^2\) and \(V\) are used {for tracking in local submaps}, registration {between them}, and loop detection to improve robustness and consistency in {long runs}. An overview of the architecture is shown in Fig. 2.
\subsection{Per-pixel uncertainty rendering}

\noindent
{\bf Intuition.} 
Before going deeper in 3D Gaussian Splatting, we first clarify the role of our per-splat appearance variance \(\sigma^2\), which is different from spatial covariance \(\Sigma\) defining its geometric extent and Spherical Harmonics (SH) coefficients defining mean appearance. In contrast, \(\sigma^2\) models uncertainty around that mean color. 
Even when the SH color correctly models the mean, view-dependent appearance, the actual color observations at a splat can vary across viewpoints. 
As illustrated in Fig. 1, larger variances appear near depth discontinuities or occlusion boundaries and in reflective or transparent regions. This is because small view changes can alter visibility and alpha weights of overlapping splats at different depths, leading to inconsistent color observations.
Moreover, we denote the appearance variance as \(\sigma^2\) to enforce positivity and follow the conventional notation of Gaussian-form uncertainty in the appearance domain. Using \(\sigma^2\) makes it clear that we optimize a true statistical variance rather than a free scale weight.
During rendering, these variances are composited into per-pixel variance $V$, allowing the renderer to quantify uncertainty directly from the splatting process. 
\\
\noindent
{\bf 3D Gaussian Splatting.} 
Followed by \cite{gaussianslam, loopy, loopsplat}, we represent the scene as a set of submaps $P^s$, each containing $N^s$ 3D Gaussians \cite{3dgs}. Each Gaussian \(G_i\) is defined by its mean position \(\mu_i \in \mathbb{R}^3\), opacity \(\alpha_i\ \in \mathbb{R}\), scales \(s_i \in \mathbb{R}^3\) and covariance matrix \(\Sigma_i \in \mathbb{R}^{3\times3}\). The corresponding color \(c_i \in \mathbb{R}^3\) is derived from spherical harmonics. We further include an additional parameter \(\sigma^2_i \in \mathbb{R}^3\) to represent the appearance variance. In summary, each submap can be described as:
\begin{equation} \label{eq1}
    P^s = \{G_i^s(\mu_i,\Sigma_i,\alpha_i,s_i,c_i,\sigma_i^2)\vert i = 1, \ldots, N^s \}
\end{equation}
With standard alpha compositing \cite{3dgs,surfacesplatting,gaussianslam}, the transmittance and weights are:
\begin{equation} \label{eq2.0}
    w_i = T_{i}\alpha_i, \quad
    T_i = \prod_j^{i-1}{(1-\alpha_j)}
\end{equation}
and the rendered color \(C\) and depth \(D\) are:
\begin{equation} \label{eq2}
    C =  \sum_i w_i c_i, \quad
    D =  \sum_i w_i z_i
\end{equation}
where \(c_i \in \mathbb{R}^3\) is the mean color and \(z_i \in \mathbb{R}\) is the camera-space depth of the projected mean of \(G_i\).

\noindent
{\bf Variance rendering.} 
For a random variable $X$ and conditioning variable $Z$, the law of total variance states:
\begin{equation} \label{eq3}
    \mathrm{Var}[X] \;=\; 
    \mathbb{E}\big[\mathrm{Var}[X \vert Z]\big] \;+\;
    \mathrm{Var}\big(\mathbb{E}[X \vert Z]\big).
\end{equation}
In our setting, $X$ denotes pixel color value, and conditioning variable $Z$ is 3D Gaussians. The per-pixel uncertainty \(\mathrm{Var}[X]\) can be then decomposed into the expected per-splat variance \(\mathbb{E}\big[\mathrm{Var}[X \vert Z]\big]\) and the variance of the splats \(\mathrm{Var}\big(\mathbb{E}[X\vert Z]\big)\). Conditioned on splat \(i\)-th, we have splat color \(c_i\) and variance \(\sigma^2_i\) can be described as:
\begin{equation} \label{eq4}
    \mathbb{E}[X \vert Z=i] = c_i, \quad
    \mathrm{Var}[X \vert Z=i] = \sigma_i^2.
\end{equation} 
Therefore, we can obtain expected per-splat variance and color by alpha blending, similar to how we compute per-pixel color in Eq. (3):
\begin{equation} \label{eq5}
    \mathbb{E}[X \vert Z] = \sum_i w_i c_i, \quad
    \mathbb{E}\big[\mathrm{Var}[X \vert Z]\big] = \sum_i w_i \sigma_i^2
\end{equation}
The variance of splats \(\mathrm{Var}\big(\mathbb{E}[X \vert Z]\big)\) is then computed using second moment of a distribution and Eq. (6):
\begin{equation}\label{eq6}
    \mathrm{Var}(\mathbb{E}[X \vert Z]) = \mathbb{E}[(X \vert Z)^2] - (\mathbb{E}[X \vert Z])^2
\end{equation}
\begin{equation} \label{eq7}
    \mathrm{Var}(\mathbb{E}[X \vert Z]) = \sum_i w_i c_i^2 - \Big(\sum_i w_i c_i\Big)^2
\end{equation}
From Eqs (4), (6) and (8), we can compute per-pixel variance \(\mathrm{Var}[X]\) denoted as \(V\) using per-splat color \(c_i\) and variance \(\sigma_i^2\). 
\begin{equation} \label{eq9}
    V = \sum_i w_i(\sigma_i^2 + c_i^2) - \Big(\sum_i w_i c_i\Big)^2
\end{equation}
We use $V$ to optimize $\sigma_i^2$ during mapping and as per-pixel weights for tracking and submap registration. By sharing the same single-pass rasterization as color and depth, $V$  enables efficient, online, in-the-loop reliability estimation.

\subsection{Mapping}
Following \cite{loopsplat,gaussianslam}, VarSplat jointly estimates camera poses \({T_j} \in SE(3)\) and Gaussian submap parameters \(G_i(\mu_i,\Sigma_i,\alpha_i,s_i,c_i,\sigma_i^2)\) from streaming RGB-D input. Given the current estimate \(T_j\) and keyframe color \(I_j\), depth \(D_j\) images, we differentiably {render the corresponding view from the current submap} to obtain the predicted color \(\hat I_j\), predicted depth \(\hat D_j\), and per-pixel uncertainty map \(V_j\) using the compositing equations Eqs (3) and (9).

\noindent
{\textbf{Submap Optimization.}}
Each submap \(P^s\) covers a limited spatial region to preserve local consistency and incremental map growth.
Following \cite{gaussianslam,loopsplat}, a new submap is initialized when the camera moves beyond a spatial threshold from the current submap centroid or when accumulated tracking uncertainty passes a preset limit.
The first keyframe seeds Gaussians by backprojecting depth points, and subsequent frames expand coverage by adding Gaussians in unobserved regions or merging overlapping ones. 
After sufficient observations, submap Gaussian parameters are jointly optimized to better align appearance and geometry with their associated keyframes. 
We leverage this stage to train the per-splat appearance variance with the following loss:
\begin{equation}
    \mathcal{L}_\text{map} = \lambda_{\text{color}} \cdot \mathcal{L}_{\text{color}} + \lambda_{\text{depth}} \cdot \mathcal{L}_{\text{depth}} + \lambda_{\text{reg}} \cdot \mathcal{L}_{\text{reg}} + \lambda_{\text{var}} \cdot \mathcal{L}_{\text{var}}
\end{equation}
where \(\lambda_\text{*}\) are hyperparameters. For color supervision, we use a weighted combination of L1 and SSIM \cite{3dgs}, while depth loss is L1 between rendered and ground-truth depth. {Similar to \cite{gsslam}, we also add regularization loss to control Gaussian scales}:
\begin{equation}
    \mathcal{L}_{\text{color}} = (1-\lambda_{\text{SSIM}}) \|\hat I - I\|_1 + \lambda_{\text{SSIM}}(1-\text{SSIM}(\hat I, I))
\end{equation}
\begin{equation}
    \mathcal{L}_{\text{depth}} =\|\hat D - D\|_1, \quad \mathcal{L}_{\text{reg}} =\|\hat s - s\|_1
\end{equation}

\noindent
{\textbf{Learning Variance from Scratch.}}
{Motivated by likelihood perspective similar to ActiveNeRF \cite{actnerf}, which treats rendered rays as independent distributions, we learn per-splat variance with same Gaussian negative log-likelihood. Unlike ActiveNeRF, we optimize this variance directly in 3DGS rasterizer without relying on neural network.}
As shown in Eq. (9), the per-pixel variance $V$ is composited from both the squared mean color $c^2$ and the per-splat variance \(\sigma^2\). 
To stay consistent with the Gaussian view, we use square L2 (MSE) for variance loss $\mathcal{L}_{\text{var}}$. 
This design reflects the Gaussian negative log-likelihood, where variance represents the second-order moment of residuals. 
Using L1 (which models Laplace scale) would break this relationship, leading to inconsistent variance estimation.
\begin{equation}
    \mathcal{L}_{\text{var}} = \frac{1}{2V}\Big(\|\hat I - I\|_2^2 + \|\hat D - D\|_2^2\Big) + \log(V)
\end{equation}
Unlike previous works that rely on pretrained uncertainty predictors \cite{wildgsslam, wildgaussians}, VarSplat learns the per-splat appearance variance \(\sigma_i^2\) end-to-end as a differentiable parameter of each Gaussian. 
Gradients from both color and depth residuals update per-splat variance, so the predicted variance reflects measurement reliability and avoids overconfidence in reflective, transparent, or glossy surfaces. The derivative with respect to the predicted per-pixel variance is given by:
\begin{equation}
    \frac{\partial \mathcal{L}_{\text{var}}}{\partial V}=-\frac{\|\hat I - I\|_2^2 + \|\hat D - D\|_2^2}{2V^2} + \frac{1}{V}
\end{equation}
which, by the chain rule and Eq.(9), gives per-splat gradient:
\begin{equation}
\frac{\partial \mathcal{L}_{\text{var}}}{\partial \sigma_i^2}
=\frac{\partial \mathcal{L}_{\text{var}}}{\partial V}\,
\frac{\partial V}{\partial \sigma_i^2}
=\frac{\partial \mathcal{L}_{\text{var}}}{\partial V}\, w_i.
\end{equation}
This allows per-splat variance $\sigma_i^2$ to adapt dynamically to residual magnitudes, while per-pixel uncertainty $V$ naturally captures fluctuations in opacity and visibility. Therefore, the system can effectively observe high uncertainty at depth discontinuities and occlusion boundaries where transmittance weights $w_i$ shift abruptly. Through optimization of camera poses, Gaussian parameters, and appearance variance, each submap progressively evolves into a locally consistent, uncertainty-aware representation. 


\subsection{Downstream Pose Estimation}

\noindent
{\textbf{Weighting.}}
Given the per-splat and per-pixel variances, we leverage them as confidence measures at both the pixel and submap levels to reduce the impact of unstable regions and photometric noise during pose estimation.
Thanks to the explicit representation and the parallel optimization of poses and Gaussian parameters, we can simultaneously freeze the variance during tracking and registration, while still training it end-to-end through the mapping stage without interrupting the gradient flow.
To obtain effective confidence weights, we normalize variance across pixels within a frame and across splats within a submap using a median-centered log scaling:
\begin{equation}
\begin{aligned}
\widetilde{V} &= \underset{\Omega}{\mathrm{median}}(\log(V)), &\widetilde{w}_p &= \exp[-(\log V - \widetilde{V}) / \tau], \\
\widetilde{\sigma^2} &= \mathrm{median}(\log \sigma^2), &
\widetilde{w}_s &= \exp[-(\log \sigma^2 - \widetilde{\sigma^2}) / \tau]
\end{aligned}
\end{equation}
where \(\tau>0\) controls the sharpness of the uncertainty weighting and \(\Omega\) denotes the valid pixels after applying inlier and alpha masks. {The normalized weights \(\widetilde{w_p}\) and \(\widetilde{w_s}\) are used in tracking and loop closure modules.} Therefore, pixels or splats with larger-than-median variance receive smaller weights, while more reliable regions receive stronger supervision.

\noindent
{\textbf{Tracking.}}
Given an incoming frame \((I,D)\), the tracker estimates the current camera pose \(T_j\) with respect to the active submap. We also apply both alpha, inlier mask to stabilize tracking and denote \(\Omega\) as valid pixels after masking. Unlike depth maps, RGB images are more susceptible to viewpoint changes, low texture, and occlusions. Therefore, optimizing a pure photometric loss for pose refinement can lead to unstable gradients. To address this, we leverage the rendered per-pixel variance as an uncertainty-aware weight to adaptively constrain unreliable pixels.
\begin{equation}
    \mathcal{L}_{\text{track}} = \sum \lambda_c \Big( \widetilde{w_p} \odot \|\hat I - I\|_1\Big)  + (1-\lambda_c) \|\hat D - D\|_1 
\end{equation}
where \(0 \leq \lambda_c \leq 1\) balances the contribution between photometric and geometric losses, and \(\widetilde{w_p}\) is the normalized variance weight defined earlier. During tracking we freeze variance parameters and stop gradients through \(\widetilde{w_p}\).

\noindent
{\textbf{Loop Detection.}}
Following LoopSplat \cite{loopsplat}, each submap is represented by keyframe descriptors for loop detection. 
Instead of computing per-pixel confidence for each keyframe, we use the learned per-splat variance \(\sigma_i^2\) to derive a submap-level reliability weight that modulates similarity.
At the submap level, rather than explicitly down-weighting appearance features, we compute the ratio of opacity $\alpha$ before and after applying the variance-based weights. This ratio encodes how much reliable appearance information remains in the submap without the need to penalize each submap separately.
\begin{equation}
    r = \frac{\sum_j \widetilde{w_s}\alpha_j}{\sum_j \alpha_j}, \quad \text{sim} = \text{cross\_sim} \odot (r_q * r_{db})
\end{equation}
where $r_\text{q}$ denotes the opacity ratio of the query submap, and $r_\text{db}$ denotes the opacity ratios of the database submaps used for computing similarity scores.

\noindent
{\textbf{Registration.}}
After detecting a loop, we register the matched submaps by localizing the query keyframes in the database submap, following \cite{loopsplat}. However, instead of using the native rendering loss \cite{gaussianslam}, we apply  the variance-derived per-pixel weight \(\widetilde w_p\) to modulate photometric loss, stabilizing the estimated viewpoint transformation. Formally, the registration loss is defined as:
\begin{equation}
    \mathcal{L}_{\text{registration}} = \sum \widetilde{w_p} \odot \|\hat I - I\|_1  + \|\hat D - D\|_1 
\end{equation}

\noindent
{\textbf{Merging Submaps and Global Refinement.}}
We merge all submaps into a consistent global representation with TSDF fusion. The merged geometry is then used to initialize Gaussian centers in the global map, which are subsequently refined using the color reconstruction loss \(\mathcal{L}_\text{color}\) as in \cite{3dgs}. In this final stage, we exclude uncertainty weights to focus on appearance quality, since unstable regions are already controlled during tracking, loop detection, and registration.

\section{Experiments}

In this section, we evaluate VarSplat against existing baselines on both synthetic and real-world datasets. Moreover, we conduct ablation studies of the proposed uncertainty model, measuring its impact on tracking, registration, and loop detection. 
Implementation details and additional results are provided in the Supplementary Material.

\subsection{Experimental Setup}
\noindent
{\bf Datasets.} 
We evaluate VarSplat on four datasets: Replica \cite{replica}, TUM-RGBD \cite{tum}, ScanNet \cite{scannet}, and ScanNet++ \cite{scannetpp}.
Following prior works \cite{gaussianslam,splatam,pointslam,loopsplat}, we use the same eight sequences in Replica.
On TUM-RGBD, we evaluate five sequences used in \cite{loopsplat}. For fair comparison on ScanNet with common baselines \cite{splatam,gaussianslam,gsslam}, we report results on six scenes. 
Following \cite{loopsplat,gsslam}, we evaluate the same five scenes a-e with IDs b20a261fdf, 8b5caf3398, fb05e13ad1, 2e74812d00, and 281bc17764 on ScanNet++.


\noindent
{\bf Baselines.} 
For camera pose estimation tracking, we use the average root mean square absolute trajectory error (ATE RMSE) \cite{atermse} on keyframes. For reconstruction quality, we follow \cite{pointslam,loopsplat} to evaluate meshes with voxel size of 1cm. We compute L1 on rendered depth and the F1 score against ground truth mesh vertices as in \cite{niceslam,loopsplat}. For rendering quality, we evaluate input training views, reporting PSNR, SSIM, and LPIPS. On ScanNet++, we also report novel view synthesis on the five standard scenes for fair comparison with \cite{gaussianslam,loopsplat}. In the tables, we highlight results, with \colorbox[HTML]{b8e0bc}{\textbf{best}}, \colorbox[HTML]{fdf6b4}{second}, and \colorbox[HTML]{ffdcc0}{third}.

\noindent
{\bf Baselines.} 
We compare VarSplat with state-of-the-art methods for dense radiance-field-based SLAM. In NeRF, we evaluate GO-SLAM \cite{goslam}, NICE-SLAM \cite{niceslam}, Point-SLAM \cite{pointslam}, ESLAM \cite{eslam}, Loopy-SLAM \cite{loopy}, and Uni-SLAM \cite{unislam}. For 3D Gaussian Splatting, we evaluate MonoGS \cite{monogs}, SplaTAM \cite{splatam}, Gaussian-SLAM \cite{gaussianslam}, LoopSplat \cite{loopsplat}, and CG-SLAM \cite{cgslam}. Among these, Uni-SLAM models uncertainty via termination-probability field, and CG-SLAM incorporates depth-driven geometric uncertainty.

\subsection{Quantitative Evaluation}

\begin{table}[t]
\centering
\caption{Tracking Performance on Replica \cite{replica} (ATE RMSE $\downarrow$ [cm]). UC indicates uncertainty. The results are highlighted as \colorbox[HTML]{b8e0bc}{\textbf{best}}, \colorbox[HTML]{fdf6b4}{second} and \colorbox[HTML]{ffdcc0}{third}. VarSplat achieves the best average performance while remaining competitive across scenes. *Photo-SLAM \cite{photoslam} use ORB-SLAM3 features \cite{orbslam3} for tracking and loop closure.}

\resizebox{\columnwidth}{!}{
\begin{tabular}{lcccccccccc}
\toprule
\textbf{Method} & UC & R0 & R1 & R2 & Off0 & Off1 & Off2 & Off3 & Off4 & \textbf{Avg.} \\
\midrule
\grayrow \multicolumn{11}{l}{\textbf{Neural Implicit Fields}} \\
NICE-SLAM \cite{niceslam}   & \xmark & 0.97 & 1.31 & 1.07 & 0.88 & 1.00 & 1.06 & 1.10 & 1.13 & 1.06 \\
ESLAM \cite{eslam}        & \xmark & 0.71 & 0.70 & 0.52 & 0.57 & 0.55 & 0.58 & 0.72 & 0.63 & 0.63 \\
Point-SLAM \cite{pointslam}   & \xmark & 0.61 & 0.41 & 0.37 & 0.38 & 0.48 & 0.54 & 0.69 & 0.72 & 0.52 \\
GO-SLAM \cite{goslam}     & \xmark & 0.34 & 0.29 & 0.29 & \third{0.32} & 0.30 & 0.39 & 0.39 & 0.46 & 0.35 \\
Loopy-SLAM \cite{loopy}   & \xmark & \second{0.24} & \third{0.24} & 0.28 & \second{0.26} & 0.40 & 0.29 & 0.22 & \third{0.35} & 0.29 \\
Uni-SLAM \cite{unislam} & \cmark &
{0.49} & {0.48} & {0.40} & {0.37} & {0.36} & {0.48} & {0.56} & {0.44} & {0.45} \\
\midrule
\grayrow \multicolumn{11}{l}{\textbf{3D Gaussian Splatting}} \\
SplaTAM \cite{splatam}        & \xmark & 0.31 & 0.40 & 0.29 & 0.47 & 0.27 & \third{0.29} & 0.32 & 0.72 & 0.38 \\
MonoGS \cite{monogs}         & \xmark & 0.33 & \best{0.22} & 0.29 & 0.36 & {0.19} & \best{0.25} & \best{0.12} & 0.81 & 0.32 \\
Gaussian-SLAM \cite{gaussianslam}  & \xmark & {0.29} & 0.29 & \third{0.22} & 0.37 & {0.23} & 0.41 & 0.30 & \third{0.35} & 0.31 \\
\textasteriskcentered Photo-SLAM \cite{photoslam} & \xmark & 0.54 & 0.39 & 0.31 & 0.52 & 0.44 & 1.28 & 0.78 & 0.58 & 0.60 \\
LoopSplat \cite{loopsplat} & \xmark &
\third{0.28} & \best{0.22} & \best{0.17} & \best{0.22} & \third{0.16} & {0.49} & \third{0.20} & \second{}{0.30} & \second{0.26} \\
CG-SLAM \cite{cgslam} & \cmark & 
{0.29} & {0.27} & {0.25} & {0.33} & \best{0.14} & \second{0.28} & {0.31} & \best{0.29} & \third{0.27} \\
\textbf{VarSplat (Ours)} & \cmark &
\best{0.20} & \second{0.23} & \second{0.19} & \second{0.26} & \second{0.15} & {0.34} & \second{0.15} & \third{0.35} & \best{0.23} \\

\bottomrule
\end{tabular}
}
\end{table}

\noindent
{\bf Tracking.} 
We report camera tracking results in Tables 1 to 4. On Replica, VarSplat improves over existing studies around 10\% through high quality synthetic data limits uncertainty's impact through high quality data.
\begin{table}[t]
\centering
\caption{Tracking Performance on ScanNet++ (ATE RMSE $\downarrow$ [cm]). VarSplat achieves the highest accuracy with robustness on large motion camera.}
\resizebox{\columnwidth}{!}{
\begin{tabular}{lcccccc}
\toprule 
\textbf{Method} & a & b & c & d & e & \textbf{Avg.} \\
\midrule
\grayrow \multicolumn{7}{l}{\textbf{Neural Implicit Fields}} \\
Point-SLAM \cite{pointslam} & 246.16 & 632.99 & 830.79 & 271.42 & 574.86 & 511.24 \\
ESLAM \cite{eslam}           & 25.15  & \second{2.15} & 27.02  & 20.89  & 35.47  & 22.14 \\
GO-SLAM \cite{goslam}       & 176.28 & 145.45 & 38.74 & 85.48 & 106.47 & 110.49 \\
Loopy-SLAM \cite{loopy} & N/A & N/A & 25.16 & 234.25 & 81.48 & 113.63 \\
\midrule
\grayrow \multicolumn{7}{l}{\textbf{3D Gaussian Splatting}} \\
SplaTAM \cite{splatam}       & 1.50 & \best{0.57} & \best{0.31} & 443.10 & {1.58} & 89.41 \\
MonoGS \cite{monogs}         & 7.00 & 3.66 & 6.37 & {3.28} & 44.09 & 12.88 \\
Gaussian-SLAM \cite{gaussianslam} & \third{1.37} & 5.97 & \third{2.70} & \third{2.35} & \second{1.02} & \third{2.68} \\
LoopSplat \cite{loopsplat} & \second{1.14} & 3.16 & {3.16} & \best{1.68} & \third{0.91} & \second{2.05} \\
\textbf{VarSplat (Ours)}    & \best{1.09} & \third{2.24} & \second{2.39} & \second{1.90} & \best{0.84 }& \best{1.69} \\
\bottomrule
\end{tabular}
}
\end{table}
On ScanNet++, VarSplat improves ATE RMSE by about 18\% over the second best method and ensures robustness in long sequences where others like SplaTAM fail (443.10 cm).
\begin{table}[t]
\centering
\caption{Tracking Performance on TUM-RGBD (ATE RMSE $\downarrow$ [cm]). UC indicates uncertainty. VarSplat outperforms both 3DGS and NeRF baselines.
*Photo-SLAM \cite{photoslam} use ORB-SLAM3 features \cite{orbslam3} for tracking and loop closure.}
\resizebox{\columnwidth}{!}{
\begin{tabular}{lccccccc}
\toprule
\textbf{Method} & UC & fr1/desk & fr1/desk2 & fr1/room & fr2/xyz & fr3/off. & \textbf{Avg.} \\
\midrule
\grayrow \multicolumn{8}{l}{\textbf{Neural Implicit Fields}} \\
NICE-SLAM \cite{niceslam}     & \xmark & 4.26 & 4.99 & 34.49 & 6.19 & 3.87 & 10.76 \\
MIPS-Fusion \cite{mipsfusion} & \xmark & 3.00 & N/A  & N/A   & 1.4  & 4.6  & N/A \\
Point-SLAM \cite{pointslam}   & \xmark & 4.34 & 4.54 & 30.92 & 1.31 & 4.8  & 8.92 \\
ESLAM \cite{eslam}             & \xmark & 2.47 & 3.69 & 29.73 & \third{1.11} & 2.42 & 7.89 \\
Co-SLAM \cite{coslam}         & \xmark & 2.40 & N/A  & N/A   & 1.70 & 2.40 & N/A \\
GO-SLAM \cite{goslam}         & \xmark & \best{1.50} & N/A & \best{4.64} & \second{0.60} & \second{1.30} & N/A \\
Loopy-SLAM \cite{loopy}   & \xmark & 3.79 & \best{3.38} & 7.03 & 1.62 & 3.41 & 3.85 \\
\midrule
\grayrow \multicolumn{8}{l}{\textbf{3D Gaussian Splatting}} \\
SplaTAM \cite{splatam}         & \xmark & 3.35 & 6.54 & 11.13 & 1.24 & 5.16 & 5.48 \\
MonoGS \cite{monogs}           & \xmark & \second{1.59} & 7.03 & 8.55 & 1.44 & \third{1.49} & 4.02 \\
Gaussian-SLAM \cite{gaussianslam} & \xmark & 2.73 & 6.03 & 14.92 & 1.39 & 5.31 & 6.08 \\
\textasteriskcentered Photo-SLAM \cite{photoslam} & \xmark & 2.60 & N/A & N/A & \best{0.35} & \best{1.00} & N/A \\
LoopSplat \cite{loopsplat} & \xmark & 2.08 & \third{3.54} & \third{6.24} & 1.58 & 3.22 & \second{3.33} \\
CG-SLAM \cite{cgslam} & \cmark & 2.43 & 4.54 & 9.39 & 1.20 & 2.45 & \third{4.0} \\
\textbf{VarSplat (Ours)}      & \cmark & \third{1.80} & \second{3.40} & \second{6.05} & 1.41 & 3.36 & \best{3.20} \\
\bottomrule
\end{tabular}
}
\end{table}
Despite lower image quality and incomplete depth on TUM-RGBD, VarSplat stabilizes motion in textureless or reflective regions by guiding correspondences without manual mask.
\begin{table}[t]
\centering
\caption{Tracking Performance on ScanNet. UC indicates uncertainty. VarSplat achieves the best overall, showing robustness to noisy indoor scenes.}
\resizebox{\columnwidth}{!}{
\begin{tabular}{lcccccccc}
\toprule
\textbf{Method} & UC & 00 & 59 & 106 & 169 & 181 & 207 & \textbf{Avg.} \\
\midrule
\grayrow \multicolumn{9}{l}{\textbf{Neural Implicit Fields}} \\
Co-SLAM \cite{coslam} & \xmark & 7.1 & 11.1 & 9.4 & \best{5.9} & 11.8 & 7.1 & 8.7 \\
NICE-SLAM \cite{niceslam} & \xmark & 12.0 & 14.0 & 7.9 & 10.9 & 13.4 & \third{6.2} & 10.7 \\
ESLAM \cite{eslam} & \xmark & 7.3 & 8.5 & {7.5} & \second{6.5} & 9.0 & \second{5.7} & 7.4 \\
Point-SLAM \cite{pointslam} & \xmark & 10.2 & 10.8 & 8.7 & 7.2 & 22.2 & 14.8  & 12.3 \\
GO-SLAM \cite{goslam} & \xmark & \third{5.4} & \third{7.5} & \second{7.0} & {7.0} & \best{6.8} & {6.9} & \second{6.8} \\
Loopy-SLAM \cite{loopy} & \xmark & \best{4.2} & \third{7.5} & 8.3 & {7.5} & 10.6 & {7.9}  & {7.7} \\
Uni-SLAM \cite{unislam} & \cmark & {6.1} & {7.8} & 7.4 & \third{5.8} & 9.8 & {5.2} & \third{7.0} \\
\midrule
\grayrow \multicolumn{9}{l}{\textbf{3D Gaussian Splatting}} \\
MonoGS \cite{monogs} & \xmark & 9.8 & 32.1 & 8.9 & 10.7 & 21.8 & 7.9  & 15.2 \\
SplaTAM \cite{splatam} & \xmark & 10.1 & 17.7 & 11.7 & 7.5 & {5.6} & {7.5} & 10.0 \\
Gaussian-SLAM \cite{gaussianslam} & \xmark & 21.2 & 12.8 & 13.5 & 13.6 & 21.0 & 13.4 & 15.9 \\
LoopSplat \cite{loopsplat} & \xmark & {6.2} & \second{7.1} & \third{7.4} & 10.6 & \second{8.5} & {6.6} & {7.7} \\
CG-SLAM \cite{cgslam} & \cmark & 7.1 & \third{7.5} & 8.9 & 8.2 & 11.6 & \best{5.3} & 8.1 \\
\textbf{VarSplat (Ours)} & \cmark & \second{4.9} & \best{5.8} & \third{6.7} & 6.7 & \third{8.9} & \third{6.2} & \best{6.5} \\
\bottomrule
\end{tabular}
}
\end{table}
On ScanNet, VarSplat consistently achieves best performance against both neural implicit and 3DGS baselines. 

\begin{table}[t]
\centering
\caption{Reconstruction Performance on Replica. VarSplat achieve third-best result after Loopy-SLAM and LoopSplat, showing that variance regularization preserves mesh quality.}
\resizebox{\columnwidth}{!}{
\begin{tabular}{llccccccccc}
\toprule
\textbf{Method} & \textbf{Metric} & R0 & R1 & R2 & Off0 & Off1 & Off2 & Off3 & Off4 & \textbf{Avg.} \\
\midrule
\grayrow \multicolumn{11}{l}{\textbf{Neural Implicit Fields}} \\
NICE-SLAM \cite{niceslam} & Depth L1 & 1.81 & 1.44 & 2.04 & 1.39 & 1.76 & 8.33 & 4.99 & 2.01 & 2.97 \\
                           & F1 [\%] & 45.0 & 44.8 & 43.6 & 50.0 & 51.9 & 39.2 & 39.9 & 36.5 & 43.9 \\
ESLAM \cite{eslam} & Depth L1 & 0.97 & 1.07 & 1.28 & 0.86 & 1.26 & 1.71 & 1.43 & 1.06 & 1.18 \\
                   & F1 [\%] & 81.0 & 82.2 & 83.9 & 78.4 & 75.5 & 77.1 & 75.9 & 79.1 & 79.1 \\
Point-SLAM \cite{pointslam} & Depth L1 & 0.53 & \second{0.22} & \second{0.46} & \second{0.30} & 0.57 & \best{0.49} & \second{0.51} & 0.64 & 0.46 \\
                             & F1 [\%] & 86.9 & \second{92.3} & \third{90.8} & \second{93.8} & \best{91.6} & \best{89.0} & \third{88.2} & 85.6 & 89.8 \\
Loopy-SLAM \cite{loopy} & Depth L1 & \best{0.30} & \best{0.20} & \best{0.42} & \best{0.23} & \best{0.46} & \second{0.60} & \best{0.37} & \best{0.24} & \best{0.35} \\
                              & F1 [\%] & \best{91.6} & \best{92.4} & {90.6} & \best{93.9} & \best{91.6} & {88.5} & \best{89.0} & \best{88.7} & \best{90.8} \\
\midrule
\grayrow \multicolumn{11}{l}{\textbf{3D Gaussian Splatting}} \\
SplaTAM \cite{splatam} & Depth L1 & \third{0.43} & 0.38 & 0.54 & 0.44 & 0.66 & 1.05 & 1.60 & 0.68 & 0.72 \\
                        & F1 [\%] & 89.3 & 88.2 & 88.0 & 91.7 & 90.0 & 85.1 & 77.1 & 80.1 & 86.2 \\
Gaussian-SLAM \cite{gaussianslam} & Depth L1 & 0.61 & 0.25 & 0.54 & 0.50 & {0.52} & 0.98 & 1.63 & {0.42} & 0.68 \\
                                  & F1 [\%] & 88.8 & 91.4 & 90.5 & 91.0 & 87.3 & 84.2 & 87.4 & 88.9 & 88.7 \\
LoopSplat \cite{loopsplat} & Depth L1 & \second{0.39} & \third{0.23} & {0.52} & \third{0.32} & \third{0.51} & {0.63} & {1.09} & \third{0.40} & \third{0.51} \\
                          & F1 [\%] & \third{90.6} & \third{91.9} & \best{91.1} & \third{93.3} & \third{90.4} & \second{88.9} & \second{88.7} & \second{88.3} & \second{90.4} \\
\textbf{VarSplat (Ours)} & Depth L1 & \second{0.39} & 0.26 & \third{0.51} & 0.33 & \best{0.46} & \second{0.60} & \third{1.08} & \second{0.35} & \second{0.50} \\
                          & F1 [\%] & \second{90.7} & 91.8 & \second{91.0} & \third{93.3} & 90.1 & \third{88.7} & 87.7 & \second{88.3} & \third{90.2} \\
\bottomrule
\end{tabular}
}
\end{table}
\noindent
{\bf Reconstruction.} 
As shown in Table 5, Depth L1 decreases slightly (0.50 vs. 0.51 for LoopSplat) and F1 is essentially unchanged (90.2 vs. 90.4), indicating tighter alignment without surface inflation. As noted in LoopSplat \cite{loopsplat}, Loopy-SLAM \cite{loopy} and Point-SLAM \cite{pointslam} sample points using ground truth depth, which assumes perfect depth input. These results also show that using the per-pixel uncertainty map to regularize the photometric loss does not degrade mesh reconstruction quality.

\begin{table}[t]
\centering
\caption{Rendering performance on 3 datasets. VarSplat
achieves competitive results on both synthetic and real-world datasets. \textcolor{lightgray}{Gray} indicates evaluation on submaps rather than global map.}
\resizebox{\columnwidth}{!}{
\begin{tabular}{lccccccccc}
\toprule
\textbf{Dataset} & \multicolumn{3}{c}{\textbf{Replica} \cite{replica}} & \multicolumn{3}{c}{\textbf{TUM} \cite{tum}} & \multicolumn{3}{c}{\textbf{ScanNet} \cite{scannet}} \\
\cmidrule(lr){2-4} \cmidrule(lr){5-7} \cmidrule(lr){8-10}
\textbf{Method} & PSNR $\uparrow$ & SSIM $\uparrow$ & LPIPS $\downarrow$ & PSNR $\uparrow$ & SSIM $\uparrow$ & LPIPS $\downarrow$ & PSNR $\uparrow$ & SSIM $\uparrow$ & LPIPS $\downarrow$ \\
\midrule
NICE-SLAM \cite{niceslam} & 24.42 & 0.892 & 0.233 & 14.86 & 0.614 & 0.441 & 17.54 & 0.621 & 0.548 \\
ESLAM~\cite{eslam} & 28.06 & 0.923 & 0.245 & 15.26 & 0.478 & 0.569 & 15.29 & 0.658 & {0.488} \\
Point-SLAM \cite{pointslam} & 35.17 & {0.975} & {0.124} & 16.62 & 0.696 & 0.526 & \second{19.82} & 0.751 & 0.514 \\
Loopy-SLAM \cite{loopy} & \third{35.47} & \third{0.981} & \second{0.109} & 12.94 & 0.489 & 0.645 & 15.23 & 0.629 & 0.671 \\
SplaTAM \cite{splatam} & 34.11 & 0.970 & \best{0.100} & \second{22.80} & \best{0.893} & \best{0.178} & \third{19.14} & 0.716 & \best{0.358} \\
\textcolor{lightgray}{Gaussian-SLAM \cite{gaussianslam} } & \textcolor{lightgray}{42.08} & \textcolor{lightgray}{0.996} & \textcolor{lightgray}{0.018} & \textcolor{lightgray}{25.05} & \textcolor{lightgray}{0.929} & \textcolor{lightgray}{0.168} & \textcolor{lightgray}{27.67} & \textcolor{lightgray}{0.923} & \textcolor{lightgray}{0.248} \\
LoopSplat \cite{loopsplat} & \second{36.63} & \second{0.985} & \third{0.112} & \third{22.72} & \third{0.873} & \third{0.259} & \best{24.92} & \second{0.845} & \third{0.425} \\
VarSplat (Ours) & \best{37.15} & \best{0.986} & \second{0.109} & \best{23.14} & \second{0.883} & \second{0.248} & \best{24.92} & \best{0.848} & \second{0.422} \\
\bottomrule
\end{tabular}
}
\end{table}
\begin{table}[t]
\centering
\caption{Novel View Synthesis on ScanNet++ (PSNR $\uparrow$). VarSplat achives the best result.}
\resizebox{0.95\columnwidth}{!}{
\begin{tabular}{lcccccc}
\toprule
\textbf{Method} & a & b & c & d & e & \textbf{Avg.} \\
\midrule
ESLAM \cite{eslam}          & 13.63 & 11.86 & 11.83 & 10.59 & 10.64 & 11.71 \\
SplaTAM \cite{splatam}      & 23.95 & 22.66 & 13.95 & 8.47  & 20.06 & 17.82 \\
Gaussian-SLAM \cite{gaussianslam} & \second{26.66} & \second{24.42} & \third{15.01} & \third{18.35} & \second{21.91} & \third{21.27} \\
LoopSplat \cite{loopsplat}   & \third{25.60} & \third{23.65} & \best{15.87} & \best{18.86} & \best{22.51} & \second{21.30} \\
\textbf{VarSplat (Ours)}   & \best{26.97} & \best{24.52} & \second{15.19} & \second{18.07} & \third{21.92} & \best{21.33} \\
\bottomrule
\end{tabular}
}
\end{table}
\begin{table}[t]
    \centering
    \caption{Effect of uncertainty on pose estimation.}
    \resizebox{0.9\columnwidth}{!}{
    \begin{tabular}{ccccc}
    \toprule
        Model & Tracking & Loop  & Registration & ATE RMSE $\downarrow$\\
    \midrule
        I & \xmark & \xmark & \xmark & 8.20 \\
        II & \cmark  & \xmark & \xmark & 7.63 \\
        III & \cmark & \cmark & \xmark & \second{7.49}\\
        IV & \xmark & \cmark & \cmark & \third{7.51}\\
        V & \cmark & \cmark & \cmark & \best{6.53} \\
    \bottomrule
    \end{tabular}
    }
\end{table}

\begin{table}[t]
    \centering
    \caption{Ablation on Variance Training.}
    \resizebox{0.9\columnwidth}{!}{
    \begin{tabular}{cccccc}
    \toprule
        Model & Track Frozen & Depth & Square L2  & ATE RMSE $\downarrow$\\
    \midrule
        I &  \xmark & \cmark  & \cmark & 7.55 \\
        II & \cmark & \xmark & \cmark &  \second{7.17} \\
        II & \cmark & \cmark & \xmark &  \third{7.38} \\
        IV & \cmark & \cmark & \cmark  & \best{6.53}\\
    \bottomrule
    \end{tabular}
    }
\end{table}

\begin{table}
\centering
\caption{Runtime on Replica/Room0 using A100 80GB. Per-frame runtime is computed as total optimization time divided by the sequence length.}
\resizebox{\columnwidth}{!}{
\begin{tabular}{lcccccc}
\toprule
\textbf{Method} & Mapping & Mapping & Tracking & Tracking & ATE  \\
 & /fr(s) & /iter(ms) & /fr(s) & /iter(ms) & RMSE \\
\midrule
NICE-SLAM \cite{niceslam}          & \third{5.8} & \third{90.6} & 8.1 & \best{20.8} & 0.97 \\
Point-SLAM \cite{pointslam} & 28.7 & 93.1 & 7.1 & \second{29.0} & 0.61 \\
SplaTAM \cite{splatam}      & \third{5.8} & 96.8 & \third{3.5} & 86.8  & \third{0.31} \\
LoopSplat \cite{loopsplat}   & \best{1.2} & \best{30.1} & \best{1.8} & \third{29.3} & \second{0.28} \\
\textbf{VarSplat (Ours)}   & \second{1.9} & \second{50.3} & \second{2.0} & 34.1 & \best{0.20} \\
\bottomrule
\end{tabular}
}
\end{table}

\noindent
{\bf Rendering.} 
In Table 6, we evaluate rendering quality on input views for Replica, TUM-RGBD, and ScanNet, and in Table 7 we report novel view synthesis on ScanNet++. VarSplat consistently achieves competitive or superior rendering across all four datasets. 

\subsection{Ablation studies}
We conduct all ablation studies on six ScanNet scenes.
\begin{figure*}
  \centering
  \includegraphics[width=\textwidth]{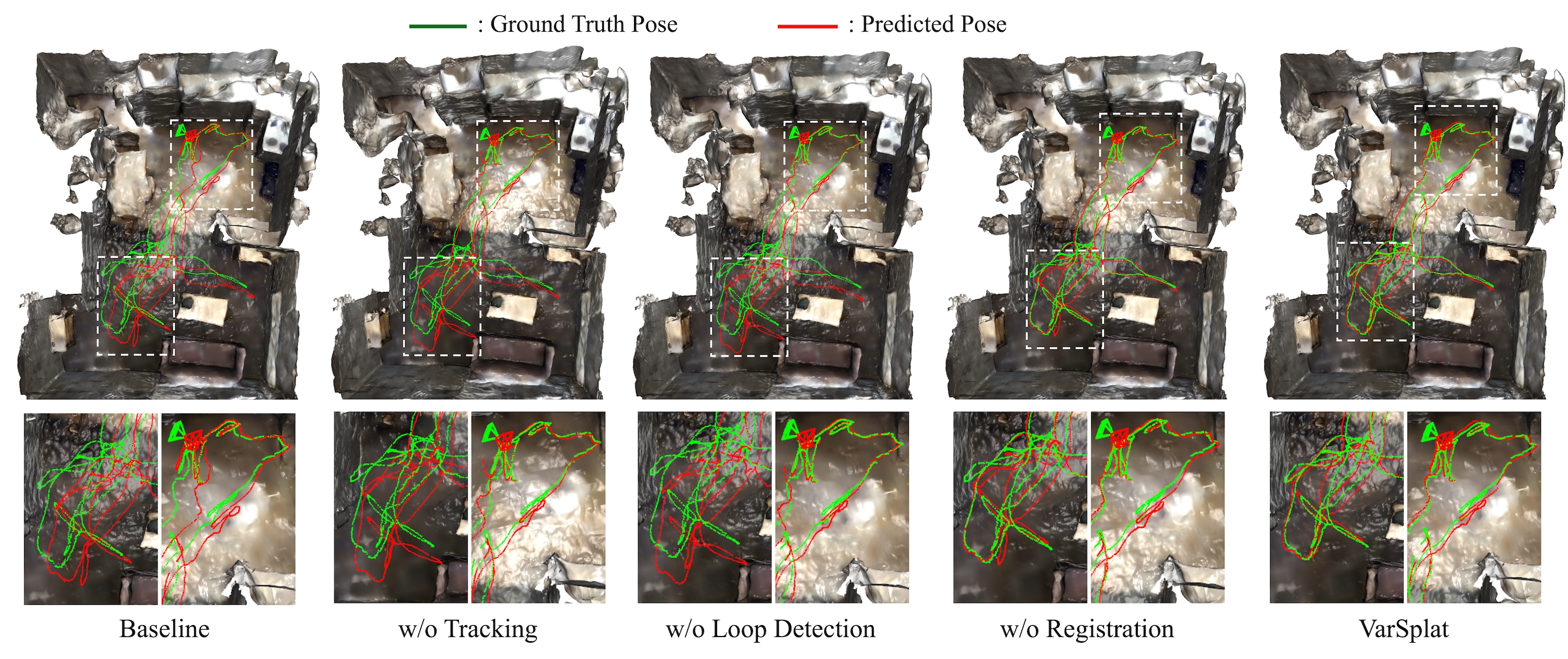}
  \caption{Uncertainty ablation on ScanNet (scene0181). Without uncertainty, tracking jitters, loop detection has long-range drift, and registration ghosts submaps. With VarSplat enabled, the trajectory is smooth and overlaps align.}
  \label{fig:short}
\end{figure*}
\begin{figure}
  \centering
  \includegraphics[width=\linewidth]{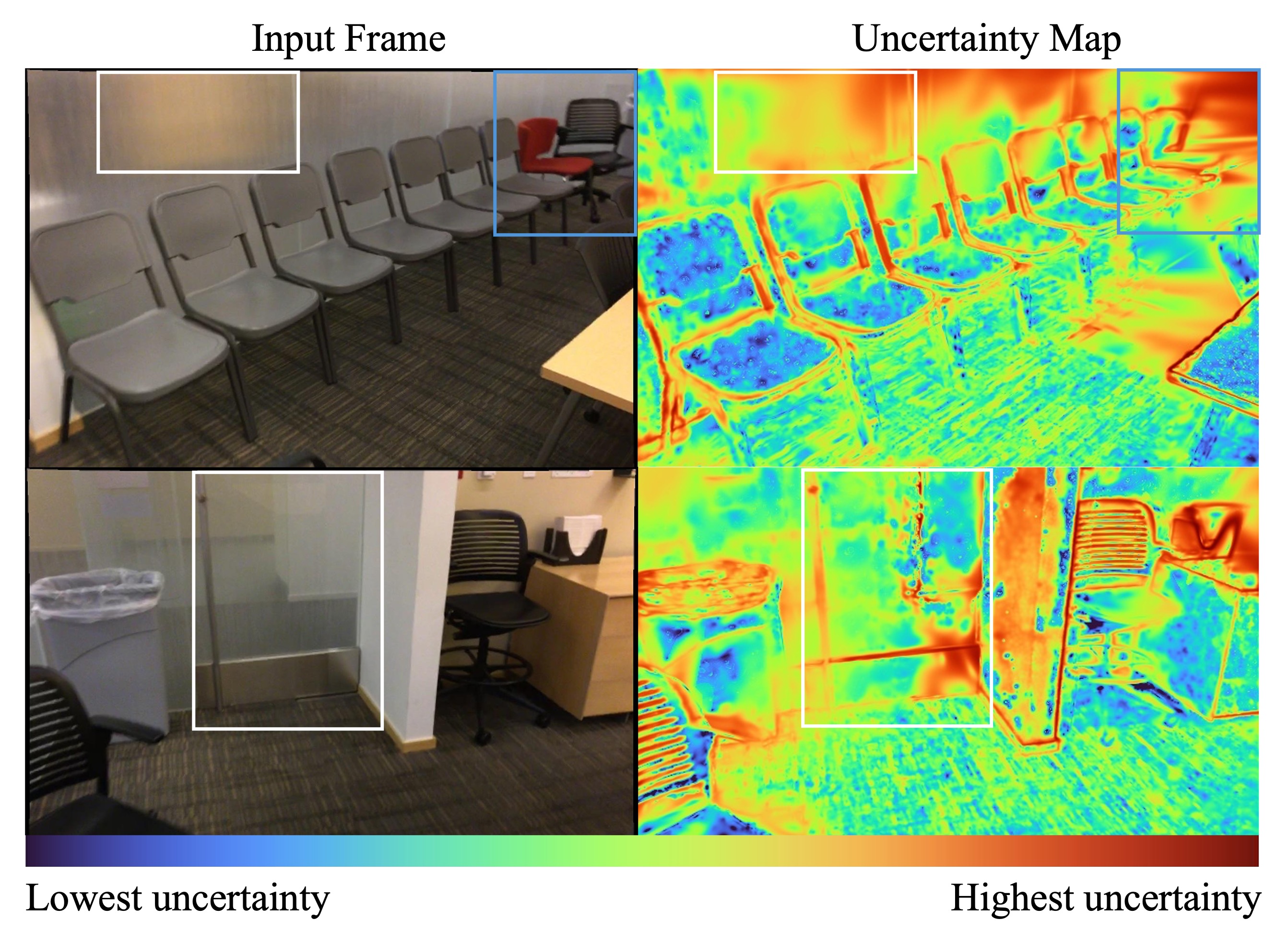}
   \caption{Visualization of challenging conditions (scene0169 ScanNet). In the top example, uncertainty is high on the texture-poor wall region (white box), while the map continues to grow (blue box). In the bottom example, the uncertainty map can distinguish reliable regions on transparent surface.}
   
\end{figure}
\begin{figure}
  \centering
  \includegraphics[width=0.95\columnwidth]{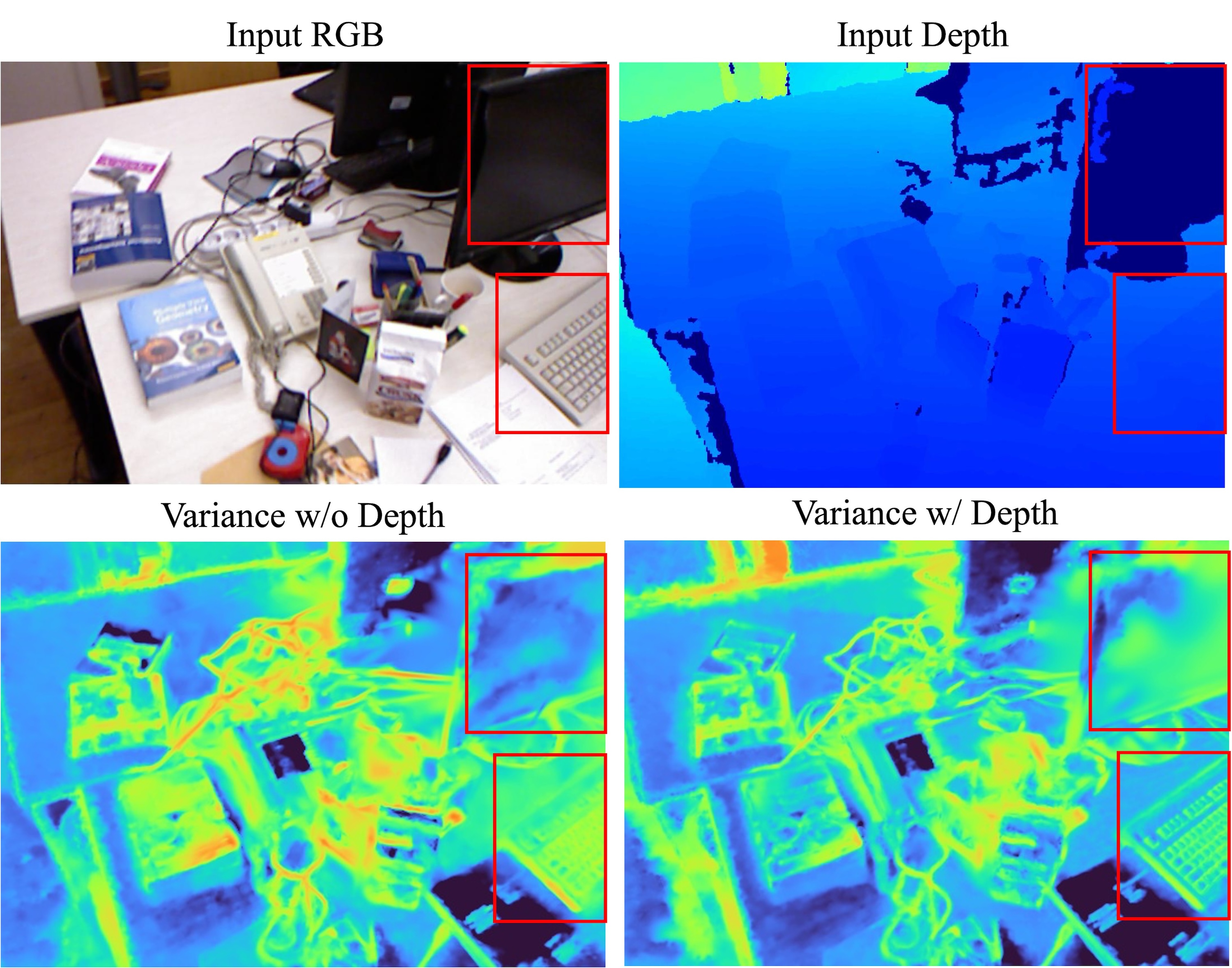}
  \caption{Per-pixel uncertainty with vs. without depth on TUM-RGBD (fr1/desk2). With depth, uncertainty focuses on textureless areas with depth holes and stays low on well constrained surfaces, avoiding overconfidence on glossy areas.}
  \label{fig:short}
\end{figure}

\noindent
{\bf Uncertainty on pose estimation.} 
Table 8 and Figure 3 present the ablation where uncertainty is applied to tracking, loop detection, and registration. Without it, pose estimation on noisy regions can cause jitter and long-range drifts. Adding uncertainty to tracking smooths local motion, in loop detection reduces false closures on repeated structure, and in registration removes overlap ghosting and tightens medium range alignment. Using all three together gives the best results, allowing VarSplat both local and global robustness and providing more reliable 3D maps.

\noindent
{\bf Variance Training.} 
Table 9 analyze learning per-splat variance with Gaussian negative log-likelihood. Freezing variance during tracking avoids conflicts with pose optimization and leads to more stable trajectories. Moreover, loop closure happens after submaps construction to check overlap. Therefore, gradients do not reach variance at this stage, so it remain frozen. Adding depth residuals tie uncertainty to geometry and prevents distractor-driven variance that misaligns submaps. Using square L2 for photometric and depth residuals matches the Gaussian model assumed by the NLL and provides smoother gradients than L1.

\noindent
{\bf Runtime.} 
As shown in Table 10, VarSplat achieve competitive runtime with recent 3DGS-SLAM systems.
Learning and rendering variance raises computation cost, with corresponding improvements in tracking performance.
\section{Conclusion}


We present VarSplat, an uncertainty-aware 3DGS-SLAM system that learns per-splat appearance variance \(\sigma^2\) and, via the law of total variance, renders differentiable per-pixel uncertainty map \(V\). Both \(\sigma^2\) and \(V\) are used for tracking, loop detection, and registration in submap-based pipeline, reducing drift and stabilizing poses. Across four datasets, this integration achieves robust and competitive-to-superior performance. Limitations and future works are provided in Supplementary Material.

\paragraph{Acknowledgement.}
This project was supported by resources provided by the Office of Research Computing at George Mason University (URL: https://orc.gmu.edu) and funded in part by grants from the National Science Foundation (Award Number 2018631).

{
    \small
    \bibliographystyle{ieeenat_fullname}
    \bibliography{main}

@String(TOG= {ACM Trans. Graph.})

@String(TOG   = {ACM TOG})

@article{replica,
  title={The replica dataset: A digital replica of indoor spaces},
  author={Straub, Julian and Whelan, Thomas and Ma, Lingni and Chen, Yufan and Wijmans, Erik and Green, Simon and Engel, Jakob J and Mur-Artal, Raul and Ren, Carl and Verma, Shobhit and others},
  journal={arXiv preprint arXiv:1906.05797},
  year={2019}
}

@inproceedings{tum,
  title={A benchmark for the evaluation of RGB-D SLAM systems},
  author={Sturm, J{\"u}rgen and Engelhard, Nikolas and Endres, Felix and Burgard, Wolfram and Cremers, Daniel},
  booktitle={2012 IEEE/RSJ international conference on intelligent robots and systems},
  pages={573--580},
  year={2012},
  organization={IEEE}
}

@inproceedings{scannet,
  title={{ScanNet: Richly-annotated 3d reconstructions of indoor scenes}},
  author={Dai, Angela and Chang, Angel X and Savva, Manolis and Halber, Maciej and Funkhouser, Thomas and Nie{\ss}ner, Matthias},
  booktitle={Proceedings of the IEEE conference on computer vision and pattern recognition},
  pages={5828--5839},
  year={2017}
}

@inproceedings{scannetpp,
  title={{ScanNet++: A high-fidelity dataset of 3d indoor scenes}},
  author={Yeshwanth, Chandan and Liu, Yueh-Cheng and Nie{\ss}ner, Matthias and Dai, Angela},
  booktitle={Proceedings of the IEEE/CVF International Conference on Computer Vision},
  pages={12--22},
  year={2023}
}

@article{3dgs,
  title={3D Gaussian splatting for real-time radiance field rendering.},
  author={Kerbl, Bernhard and Kopanas, Georgios and Leimk{\"u}hler, Thomas and Drettakis, George},
  journal={ACM Trans. Graph.},
  volume={42},
  number={4},
  pages={139--1},
  year={2023}
}

@article{nerf,
  title={{NeRF: Representing scenes as neural radiance fields for view synthesis}},
  author={Mildenhall, Ben and Srinivasan, Pratul P and Tancik, Matthew and Barron, Jonathan T and Ramamoorthi, Ravi and Ng, Ren},
  journal={Communications of the ACM},
  volume={65},
  number={1},
  pages={99--106},
  year={2021},
  publisher={ACM New York, NY, USA}
}

@inproceedings{surfacesplatting,
  title={Surface splatting},
  author={Zwicker, Matthias and Pfister, Hanspeter and Van Baar, Jeroen and Gross, Markus},
  booktitle={Proceedings of the 28th annual conference on Computer graphics and interactive techniques},
  pages={371--378},
  year={2001}
}

@inproceedings{denseslam,
  title={Dense visual SLAM for RGB-D cameras},
  author={Kerl, Christian and Sturm, J{\"u}rgen and Cremers, Daniel},
  booktitle={2013 IEEE/RSJ international conference on intelligent robots and systems},
  pages={2100--2106},
  year={2013},
  organization={IEEE}
}

@inproceedings{atermse,
  title={A benchmark for the evaluation of RGB-D SLAM systems},
  author={Sturm, J{\"u}rgen and Engelhard, Nikolas and Endres, Felix and Burgard, Wolfram and Cremers, Daniel},
  booktitle={2012 IEEE/RSJ international conference on intelligent robots and systems},
  pages={573--580},
  year={2012},
  organization={IEEE}
}

@article{ssim,
  title={Image quality assessment: from error visibility to structural similarity},
  author={Wang, Zhou and Bovik, Alan C and Sheikh, Hamid R and Simoncelli, Eero P},
  journal={IEEE transactions on image processing},
  volume={13},
  number={4},
  pages={600--612},
  year={2004},
  publisher={IEEE}
}

@article{wildgaussians,
  title={{WildGaussians}: 3d gaussian splatting in the wild},
  author={Kulhanek, Jonas and Peng, Songyou and Kukelova, Zuzana and Pollefeys, Marc and Sattler, Torsten},
  journal={arXiv preprint arXiv:2407.08447},
  year={2024}
}

@article{fisherrf,
  title={{FisherRF}: Active view selection and uncertainty quantification for radiance fields using fisher information},
  author={Jiang, Wen and Lei, Boshu and Daniilidis, Kostas},
  journal={arXiv preprint arXiv:2311.17874},
  year={2023}
}

@inproceedings{bayesnerf,
  title={Bayes' Rays: Uncertainty quantification for neural radiance fields},
  author={Goli, Lily and Reading, Cody and Sell{\'a}n, Silvia and Jacobson, Alec and Tagliasacchi, Andrea},
  booktitle={Proceedings of the IEEE/CVF Conference on Computer Vision and Pattern Recognition},
  pages={20061--20070},
  year={2024}
}

@article{modelinguncertaintygs,
  title={Modeling Uncertainty for Gaussian Splatting},
  author={Aira, Luca Savant and Valsesia, Diego and Magli, Enrico},
  journal={IEEE Transactions on Neural Networks and Learning Systems},
  year={2025},
  publisher={IEEE}
}

@article{variationalgs,
  title={Variational multi-scale representation for estimating uncertainty in 3d gaussian splatting},
  author={Li, Ruiqi and Cheung, Yiu-ming},
  journal={Advances in Neural Information Processing Systems},
  volume={37},
  pages={87934--87958},
  year={2024}
}

@inproceedings{pup3dgs,
  title={{PUP 3D-GS}: Principled uncertainty pruning for 3d gaussian splatting},
  author={Hanson, Alex and Tu, Allen and Singla, Vasu and Jayawardhana, Mayuka and Zwicker, Matthias and Goldstein, Tom},
  booktitle={Proceedings of the Computer Vision and Pattern Recognition Conference},
  pages={5949--5958},
  year={2025}
}

@article{3dgsmcmc,
  title={3d gaussian splatting as markov chain monte carlo},
  author={Kheradmand, Shakiba and Rebain, Daniel and Sharma, Gopal and Sun, Weiwei and Tseng, Yang-Che and Isack, Hossam and Kar, Abhishek and Tagliasacchi, Andrea and Yi, Kwang Moo},
  journal={Advances in Neural Information Processing Systems},
  volume={37},
  pages={80965--80986},
  year={2024}
}

@inproceedings{uncertaintysemanticgs,
  title={Modeling Uncertainty in 3D Gaussian Splatting through Continuous Semantic Splatting},
  author={Wilson, Joey and Almeida, Marcelino and Sun, Min and Mahajan, Sachit and Ghaffari, Maani and Ewen, Parker and Ghasemalizadeh, Omid and Kuo, Cheng-Hao and Sen, Arnie},
  booktitle={2025 IEEE International Conference on Robotics and Automation (ICRA)},
  pages={3284--3290},
  year={2025},
  organization={IEEE}
}

@inproceedings{stochasticnerf,
  title={Stochastic neural radiance fields: Quantifying uncertainty in implicit 3d representations},
  author={Shen, Jianxiong and Ruiz, Adria and Agudo, Antonio and Moreno-Noguer, Francesc},
  booktitle={2021 International Conference on 3D Vision (3DV)},
  pages={972--981},
  year={2021},
  organization={IEEE}
}

@inproceedings{conditionalflownerf,
  title={Conditional-Flow {NeRF}: Accurate 3d modelling with reliable uncertainty quantification},
  author={Shen, Jianxiong and Agudo, Antonio and Moreno-Noguer, Francesc and Ruiz, Adria},
  booktitle={European Conference on Computer Vision},
  pages={540--557},
  year={2022},
  organization={Springer}
}

@inproceedings{actnerf,
  title={ActNeRF: Uncertainty-aware active learning of nerf-based object models for robot manipulators using visual and re-orientation actions},
  author={Dasgupta, Saptarshi and Gupta, Akshat and Tuli, Shreshth and Paul, Rohan},
  booktitle={2024 IEEE/RSJ International Conference on Intelligent Robots and Systems (IROS)},
  pages={13062--13069},
  year={2024},
  organization={IEEE}
}

@article{bayesiannerf,
  title={Bayesian {NeRF}: Quantifying Uncertainty With Volume Density for Neural Implicit Fields},
  author={Lee, Sibaek and Kang, Kyeongsu and Ha, Seongbo and Yu, Hyeonwoo},
  journal={IEEE Robotics and Automation Letters},
  year={2025},
  publisher={IEEE}
}

@article{deepfactors,
  title={{DeepFactors}: Real-time probabilistic dense monocular slam},
  author={Czarnowski, Jan and Laidlow, Tristan and Clark, Ronald and Davison, Andrew J},
  journal={IEEE Robotics and Automation Letters},
  volume={5},
  number={2},
  pages={721--728},
  year={2020},
  publisher={IEEE}
}

@inproceedings{remode,
  title={{REMODE}: Probabilistic, monocular dense reconstruction in real time},
  author={Pizzoli, Matia and Forster, Christian and Scaramuzza, Davide},
  booktitle={2014 IEEE international conference on robotics and automation (ICRA)},
  pages={2609--2616},
  year={2014},
  organization={IEEE}
}

@article{svo,
  title={{SVO}: Semidirect visual odometry for monocular and multicamera systems},
  author={Forster, Christian and Zhang, Zichao and Gassner, Michael and Werlberger, Manuel and Scaramuzza, Davide},
  journal={IEEE Transactions on Robotics},
  volume={33},
  number={2},
  pages={249--265},
  year={2016},
  publisher={IEEE}
}

@inproceedings{loopsplat,
  title={{LoopSplat}: Loop closure by registering 3d gaussian splats},
  author={Zhu, Liyuan and Li, Yue and Sandstr{\"o}m, Erik and Huang, Shengyu and Schindler, Konrad and Armeni, Iro},
  booktitle={2025 International Conference on 3D Vision (3DV)},
  pages={156--167},
  year={2025},
  organization={IEEE}
}

@inproceedings{loopy,
  title={{Loopy-SLAM}: Dense neural slam with loop closures},
  author={Liso, Lorenzo and Sandstr{\"o}m, Erik and Yugay, Vladimir and Van Gool, Luc and Oswald, Martin R},
  booktitle={Proceedings of the IEEE/CVF conference on computer vision and pattern recognition},
  pages={20363--20373},
  year={2024}
}

@inproceedings{splatam,
  title={{SplaTAM}: Splat track \& map 3d gaussians for dense rgb-d slam},
  author={Keetha, Nikhil and Karhade, Jay and Jatavallabhula, Krishna Murthy and Yang, Gengshan and Scherer, Sebastian and Ramanan, Deva and Luiten, Jonathon},
  booktitle={Proceedings of the IEEE/CVF Conference on Computer Vision and Pattern Recognition},
  pages={21357--21366},
  year={2024}
}

@inproceedings{monogs,
  title={{Gaussian Splatting SLAM}},
  author={Matsuki, Hidenobu and Murai, Riku and Kelly, Paul HJ and Davison, Andrew J},
  booktitle={Proceedings of the IEEE/CVF Conference on Computer Vision and Pattern Recognition},
  pages={18039--18048},
  year={2024}
}

@article{gaussianslam,
  title={{Gaussian-SLAM}: Photo-realistic dense slam with gaussian splatting},
  author={Yugay, Vladimir and Li, Yue and Gevers, Theo and Oswald, Martin R},
  journal={arXiv preprint arXiv:2312.10070},
  year={2023}
}

@inproceedings{goslam,
  title={{Go-SLAM}: Global optimization for consistent 3d instant reconstruction},
  author={Zhang, Youmin and Tosi, Fabio and Mattoccia, Stefano and Poggi, Matteo},
  booktitle={Proceedings of the IEEE/CVF International Conference on Computer Vision},
  pages={3727--3737},
  year={2023}
}

@inproceedings{pointslam,
  title={{Point-SLAM} Dense neural point cloud-based slam},
  author={Sandstr{\"o}m, Erik and Li, Yue and Van Gool, Luc and Oswald, Martin R},
  booktitle={Proceedings of the IEEE/CVF International Conference on Computer Vision},
  pages={18433--18444},
  year={2023}
}

@inproceedings{photoslam,
  title={{Photo-SLAM}: Real-time simultaneous localization and photorealistic mapping for monocular stereo and rgb-d cameras},
  author={Huang, Huajian and Li, Longwei and Cheng, Hui and Yeung, Sai-Kit},
  booktitle={Proceedings of the IEEE/CVF Conference on Computer Vision and Pattern Recognition},
  pages={21584--21593},
  year={2024}
}

@inproceedings{niceslam,
  title={{Nice-SLAM}: Neural implicit scalable encoding for slam},
  author={Zhu, Zihan and Peng, Songyou and Larsson, Viktor and Xu, Weiwei and Bao, Hujun and Cui, Zhaopeng and Oswald, Martin R and Pollefeys, Marc},
  booktitle={Proceedings of the IEEE/CVF conference on computer vision and pattern recognition},
  pages={12786--12796},
  year={2022}
}

@inproceedings{eslam,
  title={{ESLAM}: Efficient dense slam system based on hybrid representation of signed distance fields},
  author={Johari, Mohammad Mahdi and Carta, Camilla and Fleuret, Fran{\c{c}}ois},
  booktitle={Proceedings of the IEEE/CVF conference on computer vision and pattern recognition},
  pages={17408--17419},
  year={2023}
}

@inproceedings{coslam,
  title={{Co-SLAM}: Joint coordinate and sparse parametric encodings for neural real-time slam},
  author={Wang, Hengyi and Wang, Jingwen and Agapito, Lourdes},
  booktitle={Proceedings of the IEEE/CVF Conference on Computer Vision and Pattern Recognition},
  pages={13293--13302},
  year={2023}
}

@article{orbslam3,
  title={{ORB-SLAM3}: An accurate open-source library for visual, visual--inertial, and multimap slam},
  author={Campos, Carlos and Elvira, Richard and Rodr{\'\i}guez, Juan J G{\'o}mez and Montiel, Jos{\'e} MM and Tard{\'o}s, Juan D},
  journal={IEEE transactions on robotics},
  volume={37},
  number={6},
  pages={1874--1890},
  year={2021},
  publisher={IEEE}
}

@inproceedings{wildgsslam,
  title={{WildGS-SLAM}: Monocular gaussian splatting slam in dynamic environments},
  author={Zheng, Jianhao and Zhu, Zihan and Bieri, Valentin and Pollefeys, Marc and Peng, Songyou and Armeni, Iro},
  booktitle={Proceedings of the Computer Vision and Pattern Recognition Conference},
  pages={11461--11471},
  year={2025}
}

@article{mipsfusion,
  title={{MIPS-fusion}: Multi-implicit-submaps for scalable and robust online neural rgb-d reconstruction},
  author={Tang, Yijie and Zhang, Jiazhao and Yu, Zhinan and Wang, He and Xu, Kai},
  journal={ACM Transactions on Graphics (TOG)},
  volume={42},
  number={6},
  pages={1--16},
  year={2023},
  publisher={ACM New York, NY, USA}
}

@inproceedings{uncleslam,
  title={{UncLe-SLAM}: Uncertainty learning for dense neural slam},
  author={Sandstr{\"o}m, Erik and Ta, Kevin and Van Gool, Luc and Oswald, Martin R},
  booktitle={Proceedings of the IEEE/CVF International Conference on Computer Vision},
  pages={4537--4548},
  year={2023}
}

@inproceedings{gsslam,
  title={{GS-SLAM}: Dense visual slam with 3d gaussian splatting},
  author={Yan, Chi and Qu, Delin and Xu, Dan and Zhao, Bin and Wang, Zhigang and Wang, Dong and Li, Xuelong},
  booktitle={Proceedings of the IEEE/CVF Conference on Computer Vision and Pattern Recognition},
  pages={19595--19604},
  year={2024}
}

@inproceedings{imap,
  title={imap: Implicit mapping and positioning in real-time},
  author={Sucar, Edgar and Liu, Shikun and Ortiz, Joseph and Davison, Andrew J},
  booktitle={Proceedings of the IEEE/CVF international conference on computer vision},
  pages={6229--6238},
  year={2021}
}

@inproceedings{cgslam,
  title={{CG-SLAM}: Efficient dense rgb-d slam in a consistent uncertainty-aware 3d gaussian field},
  author={Hu, Jiarui and Chen, Xianhao and Feng, Boyin and Li, Guanglin and Yang, Liangjing and Bao, Hujun and Zhang, Guofeng and Cui, Zhaopeng},
  booktitle={European Conference on Computer Vision},
  pages={93--112},
  year={2024},
  organization={Springer}
}

@inproceedings{unislam,
  title={{Uni-SLAM}: Uncertainty-aware neural implicit slam for real-time dense indoor scene reconstruction},
  author={Wang, Shaoxiang and Xie, Yaxu and Chang, Chun-Peng and Millerdurai, Christen and Pagani, Alain and Stricker, Didier},
  booktitle={2025 IEEE/CVF Winter Conference on Applications of Computer Vision (WACV)},
  pages={2228--2239},
  year={2025},
  organization={IEEE}
}

@inproceedings{kinectfusion,
  title={{KinectFusion}: Real-time dense surface mapping and tracking},
  author={Newcombe, Richard A and Izadi, Shahram and Hilliges, Otmar and Molyneaux, David and Kim, David and Davison, Andrew J and Kohi, Pushmeet and Shotton, Jamie and Hodges, Steve and Fitzgibbon, Andrew},
  booktitle={2011 10th IEEE international symposium on mixed and augmented reality},
  pages={127--136},
  year={2011},
  organization={Ieee}
}

@inproceedings{dtam,
  title={{DTAM}: Dense tracking and mapping in real-time},
  author={Newcombe, Richard A and Lovegrove, Steven J and Davison, Andrew J},
  booktitle={2011 international conference on computer vision},
  pages={2320--2327},
  year={2011},
  organization={IEEE}
}

@inproceedings{badslam,
  title={{Bad SLAM}: Bundle adjusted direct rgb-d slam},
  author={Schops, Thomas and Sattler, Torsten and Pollefeys, Marc},
  booktitle={Proceedings of the IEEE/CVF Conference on Computer Vision and Pattern Recognition},
  pages={134--144},
  year={2019}
}

@article{bundlefusion,
  title={{BundleFusion}: Real-time globally consistent 3d reconstruction using on-the-fly surface reintegration},
  author={Dai, Angela and Nie{\ss}ner, Matthias and Zollh{\"o}fer, Michael and Izadi, Shahram and Theobalt, Christian},
  journal={ACM Transactions on Graphics (ToG)},
  volume={36},
  number={4},
  pages={1},
  year={2017},
  publisher={ACM New York, NY, USA}
}

@article{volumetricfusion,
  title={Real-time large-scale dense RGB-D SLAM with volumetric fusion},
  author={Whelan, Thomas and Kaess, Michael and Johannsson, Hordur and Fallon, Maurice and Leonard, John J and McDonald, John},
  journal={The International Journal of Robotics Research},
  volume={34},
  number={4-5},
  pages={598--626},
  year={2015},
  publisher={Sage Publications Sage UK: London, England}
}

@inproceedings{elasticfusion,
  title={{ElasticFusion}: Dense SLAM without a pose graph.},
  author={Whelan, Thomas and Leutenegger, Stefan and Salas-Moreno, Renato F and Glocker, Ben and Davison, Andrew J},
  booktitle={Robotics: science and systems},
  volume={11},
  number={3},
  year={2015},
  organization={Rome}
}

@inproceedings{pointnerf,
  title={{Point-NeRF}: Point-based neural radiance fields},
  author={Xu, Qiangeng and Xu, Zexiang and Philip, Julien and Bi, Sai and Shu, Zhixin and Sunkavalli, Kalyan and Neumann, Ulrich},
  booktitle={Proceedings of the IEEE/CVF conference on computer vision and pattern recognition},
  pages={5438--5448},
  year={2022}
}

@article{glcslam,
  title={{Glc-SLAM}: Gaussian splatting slam with efficient loop closure},
  author={Xu, Ziheng and Li, Qingfeng and Chen, Chen and Liu, Xuefeng and Niu, Jianwei},
  journal={arXiv preprint arXiv:2409.10982},
  year={2024}
}

@inproceedings{difusion,
  title={{DI-Fusion}: Online implicit 3d reconstruction with deep priors},
  author={Huang, Jiahui and Huang, Shi-Sheng and Song, Haoxuan and Hu, Shi-Min},
  booktitle={Proceedings of the IEEE/CVF Conference on Computer Vision and Pattern Recognition},
  pages={8932--8941},
  year={2021}
}

@inproceedings{voxfusion,
  title={{Vox-Fusion}: Dense tracking and mapping with voxel-based neural implicit representation},
  author={Yang, Xingrui and Li, Hai and Zhai, Hongjia and Ming, Yuhang and Liu, Yuqian and Zhang, Guofeng},
  booktitle={2022 IEEE International Symposium on Mixed and Augmented Reality (ISMAR)},
  pages={499--507},
  year={2022},
  organization={IEEE}
}

@inproceedings{park2017colored,
  title={Colored point cloud registration revisited},
  author={Park, Jaesik and Zhou, Qian-Yi and Koltun, Vladlen},
  booktitle={Proceedings of the IEEE international conference on computer vision},
  pages={143--152},
  year={2017}
}

@inproceedings{netvlad,
  title={{NetVLAD}: CNN architecture for weakly supervised place recognition},
  author={Arandjelovic, Relja and Gronat, Petr and Torii, Akihiko and Pajdla, Tomas and Sivic, Josef},
  booktitle={Proceedings of the IEEE conference on computer vision and pattern recognition},
  pages={5297--5307},
  year={2016}
}

@inproceedings{hloc,
  title={From coarse to fine: Robust hierarchical localization at large scale},
  author={Sarlin, Paul-Edouard and Cadena, Cesar and Siegwart, Roland and Dymczyk, Marcin},
  booktitle={Proceedings of the IEEE/CVF conference on computer vision and pattern recognition},
  pages={12716--12725},
  year={2019}
}
}

\clearpage
\setcounter{page}{1}
\setcounter{section}{0}
\setcounter{table}{0}
\maketitlesupplementary

\makeatletter
\renewcommand \thesection{S\@arabic\c@section}
\renewcommand\thetable{S\@arabic\c@table}
\renewcommand \thefigure{S\@arabic\c@figure}
\makeatother

In Supplementary Material, we provide implementation details with hyperparameters and per-scene results that support the conclusions in the main paper.
\begin{itemize}
    \item \textbf{Implementation.} System hardware and software details, along with changes to the rasterizer to render appearance variance.
    \item \textbf{Additional results.} Per scene PSNR, SSIM, and LPIPS on Replica, TUM RGB-D, and ScanNet
    \item \textbf{Limitations and future work.} Discussion of current limitations and directions for applying appearance variance in dynamic scenarios.
\end{itemize}

\section{Implementation.}
\label{sec:rationale}
\paragraph{System Details.}
We implement VarSplat in Python 3.10 and PyTorch 2.4.1 on NVIDIA A100 80 GB GPU with CUDA 12.6. Starting from the original 3DGS rasterizer \cite{3dgs} and its depth-rendering extension with pose \cite{gaussianslam}, we extend the renderer to propagate per-splat appearance variance via the law of total variance (Eq. 9 in the main paper) to obtain a differentiable per-pixel uncertainty map.
\paragraph{Hyperparameters.} Defaults follow LoopSplat \cite{loopsplat} for fair comparison. We report the settings used in our experiments in Table S1, including \(\lambda_c\) for tracking loss, learning rate $l_r$ for rotation and $l_t$ for translation, and optimization iterations $\text{iter}_t$ and $\text{iter}_m$ for tracking and mapping processes, $\tau$ for variance-based weighting.
Specifically, $\tau$ controls the sharpness of the uncertainty-aware weighting function. For scenes where uncertainty-based regularization is less critical or requires minimal calibration, we set $\tau = 100$ to effectively maintain nearly uniform weight distribution across pixels and splats, thereby softening the penalty on high-variance regions.
Unless noted, $\lambda_{\text{color}}, \lambda_{\text{depth}}, \lambda_{\text{reg}}$ are set to 1 and $\lambda_{\text{var}}$ is set to 0.0001 in mapping loss \(\mathcal{L}_{\text{map}}\) for all datasets.

\begin{table}[h]
    \centering
    \caption{Per-Dataset Hyperparameters.}
    \resizebox{\columnwidth}{!}{
    \begin{tabular}{cccccc}
    \toprule
        Params & Replica \cite{replica} & TUM-RGBD \cite{tum} & ScanNet \cite{scannet}  & ScanNet++ \cite{scannetpp}\\
    \midrule
        $\lambda_c$ &  0.95 & 0.6  & 0.6 & 0.5 \\
        $l_r$ & 0.0002 & 0.002 & 0.002 &  0.002 \\
        $l_t$ & 0.002 & 0.01 & 0.01 &  0.01 \\
        $\text{iter}_t$ & 60 & 200 & 200  & 300 \\
        $\text{iter}_m$ & 100 & 100 & 100  & 500\\
        $\tau$ & 10 & 50 & 5  & 10\\
    \bottomrule
    \end{tabular}
    }
\end{table}

\paragraph{Tracking loss.} 

In the tracking loss, the inlier mask \(M_{\text{inlier}}\) filters pixels whose depth residual is extreme. Concretely, a pixel is removed if its depth error exceeds 50 times the median depth error in the current frame. Pixels without valid depth are also excluded from pose optimization. For soft alpha masking \(M_\text{alpha}=\alpha^3\), we follow prior work \cite{loopsplat,gaussianslam} for loss weighting. On ScanNet++, we reinitialize the current-frame pose with ICP odometry \cite{park2017colored} whenever the tracking loss exceeds 50 times the running average.

\paragraph{Submap.} Based on motion heristics, new submap is triggered with displacement threshold $d_\text{thre} = 0.5 [m]$ and rotation threshold $\theta_\text{thre} = 50$\textdegree \cite{loopsplat}. Due to motion blur and far camera poses in ScanNet and ScanNet++, we use a different approach
for submap initialization as setting fixed iterations of 50 and 100 frames.
For new keyframe, we uniformly sample $M_k$ points to meet alpha value condition or depth discrepancy condition. $M_k$ is limited for per-dataset setting, 30k for TUM-RGBD and ScanNet, while 100K for ScaNet++, and all available points for Replica. The alpha threshold \(\alpha_\text{thre}\) is set to 0.98 for all datasets. The depth discrepancy condition is depth error passes 40 times median depth error of current frame. New Gaussians are initialized with opacity 0.5 and initial scales to the nearest neighbor. After finishing mapping optimization, we prune Gaussians based on fixed opacity threshold with 0.1 for Replica and 0.5 for remaining datasets. 

\paragraph{Loop Detection.}
We use NetVLAD \cite{netvlad} with the VGG16-NetVLAD-Pitts30K weights from HLoc \cite{hloc}, similar to \cite{loopsplat}. For each submap, we compute cosine similarities among its keyframes and define a self-similarity score $s^i_\text{self}$ as the p-th percentile of these values. We set $p=50$ on Replica, TUM-RGBD, and ScanNet, and $p=33$ on ScanNet++. 
After obtaining submap-to-submap similarities, we rescore them with the submap-level reliability ratio $r^{(s)}$ derived from per-splat variances, so $\text{sim}_k = \text{cross\_sim}_k r^{(q)} r^{(k)}$. We then filter candidates by an overlap ratio computed with the front-end poses.

\paragraph{3DGS Registration}
For each accepted loop, we select the top $k=2$ overlapping viewpoints using NetVLAD and solve a multi-view pose estimation between the two submaps \cite{loopsplat}. We optimize the relative pose and per-view exposure coefficients for the selected viewpoints. The registration objective uses the photometric loss weighted by the per-pixel variance weight \(\tilde{w_p}\) and an unweighted depth loss, as defined in the tracking and registration losses earlier. During registration, variance parameters are fixed.
\paragraph{Datasets.}
The DSLR sequences contain abrupt motions, so we evaluate only the first 250 frames of each sequence. Table S2 reports, for all evaluated scenes, the average frame to frame translation, rotation, and the frame count. Among the benchmarks, ScanNet++ shows about $10\times$ larger motion per frame than the others, which makes pose estimation more challenging and more prone to drift, highlighting the effectiveness of our approach in reducing drift for real-world SLAM.
\begin{table}[h]
    \centering
    \caption{Frame Motion on Replica \cite{replica}, TUM-RGBD \cite{tum}, ScanNet \cite{scannet}, and ScanNet++ \cite{scannetpp}.}
    \resizebox{\columnwidth}{!}{
    \begin{tabular}{cccccc}
    \toprule
        Dataset & Replica  & TUM  & ScanNet   & ScanNet++ \\
    \midrule
        Translation (cm) &  1.07 & 1.39 & 1.34 & 14.77 \\
        Rotation (\textdegree) & 0.50 & 1.37 & 0.69 & 13.43 \\
    \bottomrule
    \end{tabular}
    }
\end{table}
\section{Additional Results.}
In the main paper we report dataset level averages for rendering quality. Tables S3 to S5 list per-scene results for Replica, TUM RGB-D, and ScanNet. Across the three datasets, VarSplat is competitive on most scenes.
\section{Limitations and Future Work}
Although VarSplat improves robustness, a depth-based approach for where and when to add Gaussians limits performance when depth is sparse or missing. As seen in the per-scene TUM-RGBD results, future work should explore joint learning of appearance and geometric uncertainty with depth completion. Additionally, learning and rendering variance add computation and memory. Future work includes variance sharing across splats, pruning, and lightweight approximations of the uncertainty map.
Fianlly, Our experiments focus on mostly static scenes. Extending appearance variance to handle moving objects with variance-guided motion segmentation and dynamic mapping is a promising direction.
\begin{table}
\centering
\caption{Rendering Performance on Replica. {\scriptsize \textcolor{red}{*}} denotes evaluating on submaps instead of a global one.}
\label{tab:replica_render}
\setlength{\tabcolsep}{4pt}
\resizebox{\linewidth}{!}{
\begin{tabular}{l l rrrrrrrr r}
\toprule
Method & Metric & Rm0 & Rm1 & Rm2 & Off0 & Off1 & Off2 & Off3 & Off4 & Avg. \\
\midrule
\multirow{3}{*}{NICE-SLAM \cite{niceslam}}
  & PSNR$\uparrow$ & 22.12 & 22.47 & 24.52 & 29.07 & 30.34 & 19.66 & 22.23 & 24.94 & 24.42 \\
  & SSIM$\uparrow$ & 0.689 & 0.757 & 0.814 & 0.874 & 0.886 & 0.797 & 0.801 & 0.856 & 0.809 \\
  & LPIPS$\downarrow$ & 0.330 & 0.271 & 0.208 & 0.229 & 0.181 & 0.235 & 0.209 & 0.198 & 0.233 \\
\hdashline
\addlinespace[2pt]
\multirow{3}{*}{ESLAM \cite{eslam}}
  & PSNR$\uparrow$ & 25.25 & 27.39 & 28.09 & 30.33 & 27.04 & 27.99 & 29.27 & 29.15 & 28.06 \\
  & SSIM$\uparrow$ & 0.874 & 0.890 & 0.935 & 0.934 & 0.910 & 0.942 & 0.953 & 0.948 & 0.923 \\
  & LPIPS$\downarrow$ & 0.315 & 0.296 & 0.245 & 0.213 & 0.254 & 0.238 & 0.186 & 0.210 & 0.245 \\
\hdashline
\addlinespace[2pt]
\multirow{3}{*}{Point-SLAM \cite{pointslam}}
  & PSNR$\uparrow$ & 32.40 & {34.08} & {35.50} & 38.26 & 39.16 & 33.99 & 33.48 & 33.49 & 35.17 \\
  & SSIM$\uparrow$ & 0.974 & 0.977 & 0.982 & 0.983 & 0.986 & 0.960 & 0.960 & 0.979 & 0.975 \\
  & LPIPS$\downarrow$ & 0.113 & 0.116 & 0.111 & 0.100 & 0.118 & 0.156 & 0.132 & 0.142 & 0.124 \\
\hdashline
\addlinespace[2pt]
\multirow{3}{*}{SplaTAM \cite{splatam}}
  & PSNR$\uparrow$ & {32.86} & 33.89 & 35.25 & 38.26 & 39.17 & 31.97 & 29.70 & 31.81 & 34.11 \\
  & SSIM$\uparrow$ & 0.980 & 0.970 & 0.980 & 0.980 & 0.980 & 0.970 & 0.950 & 0.950 & 0.970 \\
  & LPIPS$\downarrow$ & 0.070 & 0.100 & 0.080 & 0.090 & 0.090 & 0.100 & 0.120 & 0.150 & 0.100 \\
\hdashline
\addlinespace[2pt]
\multirow{3}{*}{\textcolor{red}{*} Gaussian-SLAM \cite{gaussianslam}}
  & PSNR$\uparrow$ & 38.88 & 41.80 & 42.44 & 46.40 & 45.29 & 40.10 & 39.06 & 42.65 & 42.08 \\
  & SSIM$\uparrow$ & 0.993 & 0.996 & 0.996 & 0.998 & 0.997 & 0.997 & 0.997 & 0.997 & 0.996 \\
  & LPIPS$\downarrow$ & 0.017 & 0.018 & 0.019 & 0.015 & 0.016 & 0.020 & 0.020 & 0.020 & 0.018 \\
\hdashline
\addlinespace[2pt]
\multirow{3}{*}{LoopSplat \cite{loopsplat}}
  & PSNR$\uparrow$ & {33.07} & {35.32} & {36.16} & {40.82} & 40.21 & 34.67 & 35.67 & 37.10 & 36.63 \\
  & SSIM$\uparrow$ & 0.973 & 0.978 & 0.985 & 0.992 & 0.990 & 0.985 & 0.990 & 0.989 & 0.985 \\
  & LPIPS$\downarrow$ & 0.116 & 0.122 & 0.111 & 0.085 & 0.123 & 0.140 & 0.096 & 0.106 & 0.112 \\
\hdashline
\addlinespace[2pt]
\multirow{3}{*}{\textbf{VarSplat}}
  & PSNR$\uparrow$ & \best{33.93} & \best{35.82} & \best{36.50} & \best{41.06} & \best{41.12} & \best{35.61} & \best{36.05} & \best{37.49} & \best{37.16} \\
  & SSIM$\uparrow$ & \second{0.978} & \best{0.981} & \best{0.986} & \best{0.992} & \best{0.991} & \best{0.986} & \best{0.990} & \best{0.990} & \best{0.986} \\
  & LPIPS$\downarrow$ & \second{0.105} & \third{0.117} & \second{0.109} & \best{0.082} & \third{0.120} & \second{0.137} & \best{0.093} & \best{0.100} & \second{0.109} \\
\bottomrule
\end{tabular}}
\end{table}

\begin{table}
\centering
\caption{Rendering Performance on TUM RGB-D. \textcolor{red}{*} denotes evaluating on submaps instead of a global one.}
\label{tab:tum_render}
\setlength{\tabcolsep}{6pt}
\resizebox{\linewidth}{!}{
\begin{tabular}{l l rrr r}
\toprule
Method & Metric & fr1/desk & fr2/xyz & fr3/office & Avg. \\
\midrule
\multirow{3}{*}{NICE-SLAM \cite{niceslam}}
  & PSNR$\uparrow$  & 13.83 & 17.87 & 12.89 & 14.86 \\
  & SSIM$\uparrow$  & 0.569 & 0.718 & 0.554 & 0.614 \\
  & LPIPS$\downarrow$ & 0.482 & 0.344 & 0.498 & 0.441 \\
\hdashline
\addlinespace[2pt]
\multirow{3}{*}{ESLAM \cite{eslam}}
  & PSNR$\uparrow$  & 11.29 & 17.46 & 17.02 & 15.26 \\
  & SSIM$\uparrow$  & 0.666 & 0.310 & 0.457 & 0.478 \\
  & LPIPS$\downarrow$ & 0.358 & 0.698 & 0.652 & 0.569 \\
\hdashline
\addlinespace[2pt]
\multirow{3}{*}{Point-SLAM \cite{pointslam}}
  & PSNR$\uparrow$  & 13.87 & 17.56 & 18.43 & 16.62 \\
  & SSIM$\uparrow$  & 0.627 & 0.708 & 0.754 & 0.696 \\
  & LPIPS$\downarrow$ & 0.544 & 0.585 & 0.448 & 0.526 \\
\hdashline
\addlinespace[2pt]
\multirow{3}{*}{SplaTAM \cite{splatam}}
  & PSNR$\uparrow$  & 22.00 & 24.50 & 21.90 & 22.80 \\
  & SSIM$\uparrow$  & 0.857 & 0.947 & 0.876 & 0.893 \\
  & LPIPS$\downarrow$ & 0.232 & 0.100 & 0.202 & 0.178 \\
\hdashline
\addlinespace[2pt]
\multirow{3}{*}{\textcolor{red}{*} Gaussian-SLAM \cite{gaussianslam}}
  & PSNR$\uparrow$  & 24.01 & 25.02 & 26.13 & 25.05 \\
  & SSIM$\uparrow$  & 0.924 & 0.924 & 0.939 & 0.929 \\
  & LPIPS$\downarrow$ & 0.178 & 0.186 & 0.141 & 0.168 \\
\hdashline
\addlinespace[2pt]
\multirow{3}{*}{LoopSplat \cite{loopsplat}}
  & PSNR$\uparrow$  & 22.03 & 22.68 & 23.47 & 22.72 \\
  & SSIM$\uparrow$  & 0.849 & 0.892 & 0.879 & 0.873 \\
  & LPIPS$\downarrow$ & 0.307 & 0.217 & 0.253 & 0.259 \\
\hdashline
\addlinespace[2pt]
\multirow{3}{*}{\textbf{VarSplat}}
  & PSNR$\uparrow$  & \best{22.03} & \second{23.85} & \best{23.53} & \best{23.14} \\
  & SSIM$\uparrow$  & \third{0.847} & \second{0.920} & \best{0.882} & \second{0.883} \\
  & LPIPS$\downarrow$ & \third{0.311} & \second{0.189} & \second{0.248} & \second{0.248} \\
\bottomrule
\end{tabular}}
\end{table}

\begin{table}
\centering
\caption{Rendering Performance on ScanNet. \textcolor{red}{*} denotes evaluating on submaps instead of a global one.}
\label{tab:scannet_render}
\setlength{\tabcolsep}{5pt}
\resizebox{\linewidth}{!}{
\begin{tabular}{l l rrrrrr r}
\toprule
Method & Metric & 0000 & 0059 & 0106 & 0169 & 0181 & 0207 & Avg. \\
\midrule
\multirow{3}{*}{NICE-SLAM \cite{niceslam}}
  & PSNR$\uparrow$  & 18.71 & 16.55 & 17.29 & 18.75 & 15.56 & 18.38 & 17.54 \\
  & SSIM$\uparrow$  & 0.641 & 0.605 & 0.646 & 0.629 & 0.562 & 0.646 & 0.621 \\
  & LPIPS$\downarrow$ & 0.561 & 0.534 & 0.510 & 0.534 & 0.602 & 0.552 & 0.548 \\
  \hdashline
\addlinespace[2pt]
\multirow{3}{*}{ESLAM \cite{eslam}}
  & PSNR$\uparrow$  & 15.70 & 14.48 & 15.44 & 14.56 & 14.22 & 17.32 & 15.29 \\
  & SSIM$\uparrow$  & 0.687 & 0.632 & 0.628 & 0.656 & 0.696 & 0.653 & 0.658 \\
  & LPIPS$\downarrow$ & 0.449 & 0.450 & 0.529 & 0.486 & 0.482 & 0.534 & 0.488 \\
  \hdashline
\addlinespace[2pt]
\multirow{3}{*}{Point-SLAM \cite{pointslam}}
  & PSNR$\uparrow$  & 21.30 & 19.48 & 16.80 & 18.53 & 18.23 & 20.56 & 19.16 \\
  & SSIM$\uparrow$  & 0.806 & 0.765 & 0.676 & 0.688 & 0.823 & 0.750 & 0.751 \\
  & LPIPS$\downarrow$ & 0.485 & 0.499 & 0.544 & 0.542 & 0.471 & 0.544 & 0.514 \\
  \hdashline
\addlinespace[2pt]
\multirow{3}{*}{SplaTAM \cite{splatam}}
  & PSNR$\uparrow$  & 19.33 & 19.27 & 17.73 & 21.97 & 16.76 & 19.80 & 19.14 \\
  & SSIM$\uparrow$  & 0.660 & 0.792 & 0.690 & 0.776 & 0.683 & 0.696 & 0.716 \\
  & LPIPS$\downarrow$ & 0.438 & 0.289 & 0.376 & 0.281 & 0.402 & 0.341 & 0.358 \\
  \hdashline
\addlinespace[2pt]
\multirow{3}{*}{\textcolor{red}{*} Gaussian-SLAM \cite{gaussianslam}}
  & PSNR$\uparrow$  & 28.54 & 26.21 & 23.27 & 28.60 & 27.79 & 28.63 & 27.67 \\
  & SSIM$\uparrow$  & 0.926 & 0.934 & 0.926 & 0.917 & 0.923 & 0.913 & 0.923 \\
  & LPIPS$\downarrow$ & 0.271 & 0.211 & 0.217 & 0.226 & 0.277 & 0.288 & 0.248 \\
  \hdashline
\addlinespace[2pt]
\multirow{3}{*}{LoopSplat \cite{loopsplat}}
  & PSNR$\uparrow$  & 24.99 & 23.23 & 23.35 & 26.80 & 24.82 & 26.33 & 24.92 \\
  & SSIM$\uparrow$  & 0.840 & 0.831 & 0.846 & 0.877 & 0.824 & 0.854 & 0.845 \\
  & LPIPS$\downarrow$ & 0.450 & 0.400 & 0.409 & 0.346 & 0.514 & 0.430 & 0.425 \\
  \hdashline
\addlinespace[2pt]
\multirow{3}{*}{\textbf{VarSplat}}
  & PSNR$\uparrow$  & \best{25.12} & \second{23.16} & \best{23.52} & \best{26.82} & \second{24.67} & \second{26.20} & \best{24.92} \\
  & SSIM$\uparrow$  & \best{0.850} & \best{0.834} & \best{0.852} & \best{0.879} & \second{0.820} & \best{0.854} & \best{0.848} \\
  & LPIPS$\downarrow$ & \second{0.444} & \second{0.399} & \second{0.404} & \second{0.340} & 0.518 & \second{0.425} & \second{0.422} \\
\bottomrule
\end{tabular}}
\end{table}


\end{document}